\documentclass[runningheads]{llncs}

\usepackage{eccv}

\usepackage{eccvabbrv}

\usepackage{graphicx}
\usepackage{booktabs}

\usepackage[accsupp]{axessibility}  % Improves PDF readability for those with disabilities.

\usepackage{orcidlink}

\usepackage{hyperref}
\usepackage{colortbl}
\usepackage{multirow}
\usepackage{adjustbox}
\usepackage{pifont}

\usepackage{hhline}

\definecolor{tabfirst}{rgb}{1, 0.7, 0.7}
\definecolor{tabsecond}{rgb}{1, 0.85, 0.7} % orange
\definecolor{tabthird}{rgb}{1, 1, 0.7} % yellow

\begin{document}

%\title{Author Guidelines for ECCV Submission} 

%\title{Contribution-based Efficient Low-Rank Adaptation for Real Image Restoration}
\title{Contribution-based Low-Rank Adaptation with Pre-training Model for Real Image Restoration}

\titlerunning{CoLoRA with Pre-training Model for Real Image Restoration}

% \author{Dongwon Park\inst{1}\orcidlink{0000-0001-6060-9705} \and
% Hayeon Kim\inst{2}\orcidlink{0009-0008-7461-9750} \and
% Se Young Chun\inst{1,2}\thanks{Corresponding author.}\orcidlink{0000-0001-8739-8960}}

\author{Dongwon Park\inst{1} \and
Hayeon Kim\inst{2} \and
Se Young Chun\inst{1,2}\thanks{Corresponding author.}}

\authorrunning{D. Park et al.}

\institute{$^1$IPAI \& INMC, \ $^2$Dept. of Electrical and Computer Engineering, \\
Seoul National University, \ Republic of Korea,\\
\email{\{dong1park, khy5630, sychun\}@snu.ac.kr}  
}

\maketitle

\begin{abstract}
Recently, pre-trained model and efficient parameter tuning have achieved remarkable success in natural language processing and high-level computer vision with the aid of masked modeling and prompt tuning. In low-level computer vision, however, there have been limited investigations on pre-trained models and even efficient fine-tuning strategy has not yet been explored despite its importance and benefit in various real-world tasks such as alleviating memory inflation issue when integrating new tasks on AI edge devices. Here, we propose a novel efficient parameter tuning approach dubbed contribution-based low-rank adaptation (CoLoRA) for multiple image restorations along with effective pre-training method with random order degradations (PROD). Unlike prior arts that tune all network parameters, our CoLoRA effectively fine-tunes small amount of parameters by leveraging LoRA (low-rank adaptation) for each new vision task with our contribution-based method to adaptively determine layer by layer capacity for that task to yield comparable performance to full tuning. Furthermore, our PROD strategy allows to extend the capability of pre-trained models with improved performance as well as robustness to bridge synthetic pre-training and real-world fine-tuning. Our CoLoRA with PROD has demonstrated its superior performance in various image restoration tasks across diverse degradation types on both synthetic and real-world datasets for known and novel tasks. 
Project page: \url{https://janeyeon.github.io/colora/}.
\keywords{Efficient fine-tuing \and Low-rank adaptation \and  Pre-training}
\end{abstract}

\section{Introduction}
\label{sec:intro}

Image restoration (IR) is a fundamental low-level computer vision task that aims to recover the original clean image from the input data that was degraded by noise~\cite{zhang2017beyond,lehtinen2018noise2noise,zhussip2019extending,huang2021neighbor2neighbor}, blur~\cite{lai2016comparative,nah2017deep,kupyn2018deblurgan,whang2022deblurring,rim2022realistic}, and / or bad weather conditions~\cite{li2017aod,zhang2018densely,zheng2021ultra,wu2021contrastive}.
It does not only enhance the visual quality of images, but also improves the performance of mid to high-level vision downstream tasks such as classification~\cite{he2016deep,krizhevsky2017imagenet}, object detection~\cite{ren2015faster,redmon2016you}, and autonomous driving~\cite{bojarski2016end,luo2021self}.
However, existing IR methods require expensive training data per degradation and individual training of a separate model from scratch per restoration task.

Natural language processing (NLP) and high-level computer vision also have similar problems of high cost data collection and demanding training for diverse tasks. However, large-scale pre-trained models~\cite{devlin2018bert,brown2020language,pathak2017learning,gidaris2018unsupervised} and efficient parameter tuning methods~\cite{jiang2019smart, zaken2021bitfit, ding2023parameter, pfeiffer2020adapterhub,pfeiffer2020adapterfusion,houlsby2019parameter,zaken2021bitfit,cai2020tinytl,hu2021lora} have been recently proposed to address this issue for effectively adapting to new tasks.
Large-scale pre-training models are constructed, they resolve the data scarcity problem, reduce training time, and improve generalization performance through powerful learning capability and scalability for diverse tasks. 
Moreover, the efficient parameter tuning methods adjust only a small number of parameters during fine-tuning for each task, so they can significantly reduce memory and storage cost.
In IR, efficient parameter tuning has not been investigated yet, but pre-training approaches for IR~\cite{chen2021pre,li2021efficient,chen2023activating,liu2023degae} have recently emerged as a noteworthy development.

\begin{figure}[!t]
    \centering
\includegraphics[width=0.9\textwidth]{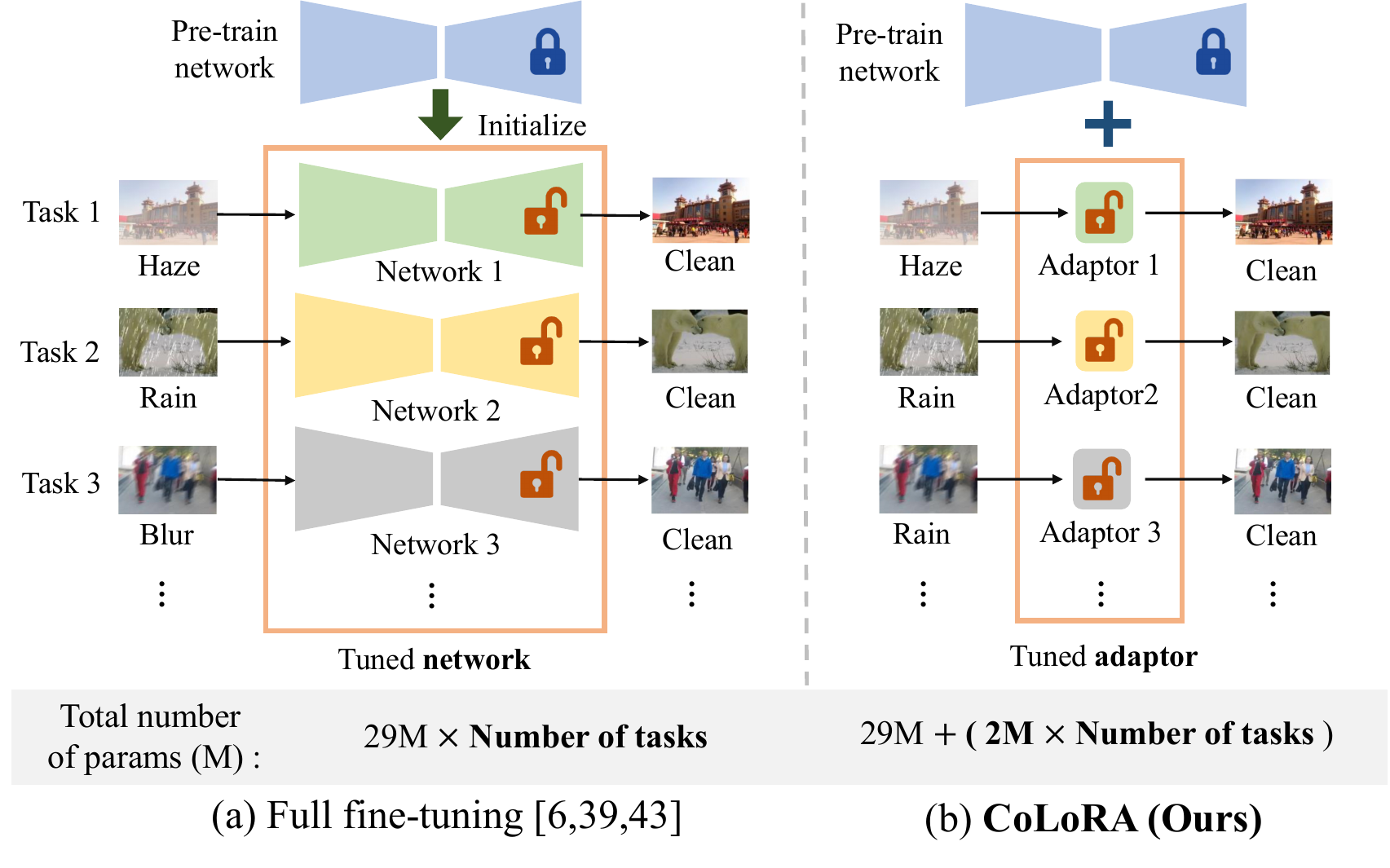}
    \caption{Illustrations of tuning strategies for novel image restoration tasks. 
    (a) Existing strategies~\cite{chen2021pre,li2021efficient,liu2023degae} for fully fine-tuning a pre-trained model for a new task.
    (b) Our proposed CoLoRA method enables parameter-efficient fine-tuning by freezing the pre-trained model and adjusting the additional adapter for novel image restoration tasks.}
    \label{Fig1_overview}
\end{figure}

The existing pre-training approaches~\cite{chen2021pre,li2021efficient,liu2023degae} for IR usually construct synthetic training data with known multiple degradations (called synthetic degradation functions), and then fine-tune the full network parameters to the real world data with novel degradation in Fig.~\ref{Fig1_overview}(a). 
This pre-training technique offers the benefit of optimizing performance using limited data in real  IR tasks where obtaining a real-world dataset containing pairs of degraded and clean images can be challenging.
However,  these methods have the disadvantage of requiring a large number of network parameters and storage memory as the number of tasks increases because the entire network parameters must be trained for new tasks, as illustrated in Fig.~\ref{Fig1_overview}.
Therefore, when various functions exist, operation may be difficult in an intelligent edge device with limited computing capacity.

In this paper, we propose an efficient parameter tuning method using a new adapter called contribution-based low-rank adaptation (CoLoRA) with a Pre-training with Random Order Degradation (PROD).  
These approaches enable the pre-trained network to adapt to IR tasks with novel degradations, as depicted in Fig.~\ref{Fig1_overview}(b).
Unlike previous works that fine-tune the entire pre-trained network, our proposed method introduces adapters for efficient parameter tuning in novel IR tasks.
While prior works must have a separate full-size fine-tuned network for each novel IR task, our proposed method only need to store the small tunable adaptor for each task (approximately 7\% of the entire network parameters for comparable performance to full-size tuning), so that it will be advantageous in terms of memory for multiple target tasks. 
The inefficiency with full fine-tuning is especially severe in modern network architectures for IR such as Restormer~\cite{zamir2022restormer} and IPT~\cite{chen2021pre}.
Our proposed PROD strengthens the pre-train model by generating low-quality training images using synthetic degradation functions and their random combinations. 
This is achieved by synergistically leveraging both random single degradations~\cite{chen2021pre,li2021efficient} and deterministic multiple degradations~\cite{liu2023degae} so that the pre-text task can be expanded for adapting IR to novel real-world complex degradation. 
Here is the summary of our contributions: 
\begin{itemize}
    \item We propose a novel efficient parameter tuning method with pre-training for IR, dubbed CoLoRA with PROD, using synthetic degradation functions for pre-training and small amount of real data for parameter tuning. 
    Our proposed method demonstrates flexible enough to work with diverse network architectures like CNNs and vision transformers, achieving state-of-the-art performance on 6 IR tasks with \emph{real} data.
    \item Our CoLoRA allows efficient parameter tuning by freezing the pre-trained model and tuning the additional adaptor with adaptive layer by layer capacity per task, still achieving state-of-the-art performance on \emph{real} 6 IR tasks.
    \item Our PROD has trained an excellent pre-training network with synthetic data that outperformed other state-of-the-art methods on 6 IR tasks with \emph{real} data after full fine-tuning.
\end{itemize}

\section{Related works}
\label{sec:rel}
\subsection{NLP and High-level Vision Tasks}

\textbf{Pre-training.} 
In NLP, self-supervised pre-training utilizing large models and billions of data~\cite{devlin2018bert,brown2020language,radford2019language} is a fundamental option.
These methods train the model to predict missing content by hiding part of the input sequence.
Various self-supervised pre-training methodologies~\cite{doersch2015unsupervised,gidaris2018unsupervised} 
have also been proposed in high-level computer vision fields.
Recently, contrastive learning~\cite{chen2020simple,he2020momentum} and transformer-based masked autoencoder methods~\cite{he2022masked,xie2022simmim} have emerged, enhancing semantic information learning. 
These pre-trained networks provide better generalization performance in various downstream tasks through efficient representation learning and extensive data utilization, even with less training data.

\noindent\textbf{Parameter efficient fine-tuning.}
To adapt the aforementioned pre-trained models to various downstream tasks, a fine-tuning process is necessary.
However, a major drawback of fine-tuning is that it comes with long training time~\cite{yang2022gram} and memory discomfort~\cite{bulo2018place}, as the learning parameters for the new task are the same as the pre-trained model.
To alleviate this, the topic of parameter efficient fine-tuning methods have been actively researched in the NLP~\cite{jiang2019smart, zaken2021bitfit, ding2023parameter, pfeiffer2020adapterhub} and high-level computer vision~\cite{jia2022visual, zhou2022learning, zhang2020side} fields.
These methodologies propose adjusting only specific parameters of the model, such as the adaptor~\cite{pfeiffer2020adapterfusion,houlsby2019parameter}, biases~\cite{zaken2021bitfit,cai2020tinytl}, prompts~\cite{jia2022visual, li2021prefix, khattak2023maple}, etc.
Recently in image generation tasks~\cite{rombach2022high,luo2023lcm,podell2023sdxl,zhang2023adding}, ControlNet~\cite{zhang2023adding} and LoRA~\cite{hu2021lora}  have been proposed. 
The ControlNet~\cite{zhang2023adding} is a method to adapt to new tasks by additionally using parameters from the decoder and middle block. 
However, this method has a disadvantage of increasing inference time. On the other hand, LoRA does not increase inference time by fixing all weights of pre-trained models and efficiently adjusting only learnable rank decomposition matrix parameters for downstream tasks.
Efficient parameter tuning has not been studied in IR, but recently pre-training methods~\cite{chen2021pre,li2021efficient,chen2023activating,liu2023degae} for IR have begun to be proposed.
We propose a novel efficient parameter tuning method, CoLoRA with PROD method specifically designed for IR tasks.

\subsection{Low-Level Vision Tasks}
\textbf{Image restoration for multiple degradations.}
Recently, methodologies~\cite{zamir2021multi,chu2021improving,wang2022uformer,zamir2022restormer,tu2022maxim,mou2022deep,chen2022simple} have been proposed that use a single network architecture to build multiple independent restoration models, each trained on different degradation datasets, demonstrating high performance in various degradation tasks. 
%This methodology demonstrated high performance in various degradation tasks.
However, these methods require numerous network parameters as it necessitates an independent network trained for each degradation tasks.
To address  this, an all-in-one IR methods~\cite{chen2022learning,li2020all,li2022all,zhang2023ingredient,luo2023controlling,valanarasu2022transweather,liu2022tape,park2023all,li2020all,park2023all,zhu2023learning} have been proposed.
Firstly, there was a method~\cite{chen2022learning} of learning a unified network for various degradations through knowledge distillation techniques.
Secondly, there were methods~\cite{li2022all,zhang2023ingredient,luo2023controlling,valanarasu2022transweather,liu2022tape} of using an adaptor to enable the unified network to adapt to various degradations.
Lastly, there were methods~\cite{li2020all,park2023all,zhu2023learning} of using an additional module corresponding to a specific degradation in a unified model,  including a classifier that selects the additional module.
All-in-one IR methods achieved high performance in various degradation tasks.
However, these methods have the limitation of requiring to re-train the unified model and adaptor or classifier to extend to new tasks.
Furthermore, all-in-one methods train from scratch without any pre-trained model. 

\noindent\textbf{Pre-training with synthetic data.}
Constructing a pre-trained model for IR tasks requires a significant quantity of low-quality images paired  with their high-quality counterparts~\cite{ledig2017photo, zhang2018ffdnet}. 
However, obtaining such image pairs in the real world presents considerable cost and difficulty~\cite{liu2023degae}.
To deal with such problems, pre-training methods using synthetic degradation functions have been proposed as illustrated in Fig.~\ref{Fig1_overview}.
IPT~\cite{chen2021pre} and EDT~\cite{li2021efficient} proposed  methods that select single degradation from multiple synthetic degradation functions such as super-resolution, Gaussian noise, and rain, thus generate low-quality images.
HAT~\cite{chen2023activating} focuses on selecting a single degradation from down-scale degradation functions. 
DegAE~\cite{liu2023degae} employs pre-determined Gaussian blur, noise, and JPEG degradation functions in a fixed order and introduces a degradation autoencoder to integrate features.
However, DegAE~\cite{liu2023degae} does not utilize global residual learning to aggregate features into a single representation. 
This omission may lead to a decrease in performance, as the original network's capabilities are not fully utilized.
These methods may not be well-suited for addressing real-world and complex degradation tasks because their limited representation during pre-training. 
Moreover, there exists a significant constraint that the entire network must be fine-tuned for each new task.
Consequently, there arises a challenge of storing and deploying distinct copies of base parameters for individual tasks, induces the issue of cost and memory.
To alleviate it, we propose novel CoLoRA, an efficient fine-tuning methodology, and PROD, an effective pre-training method.

\section{Method}
\label{sec:method}
\begin{figure*}[!b]
    % \centering
%    \hspace{-0.5cm}
    \includegraphics[width=1\textwidth]{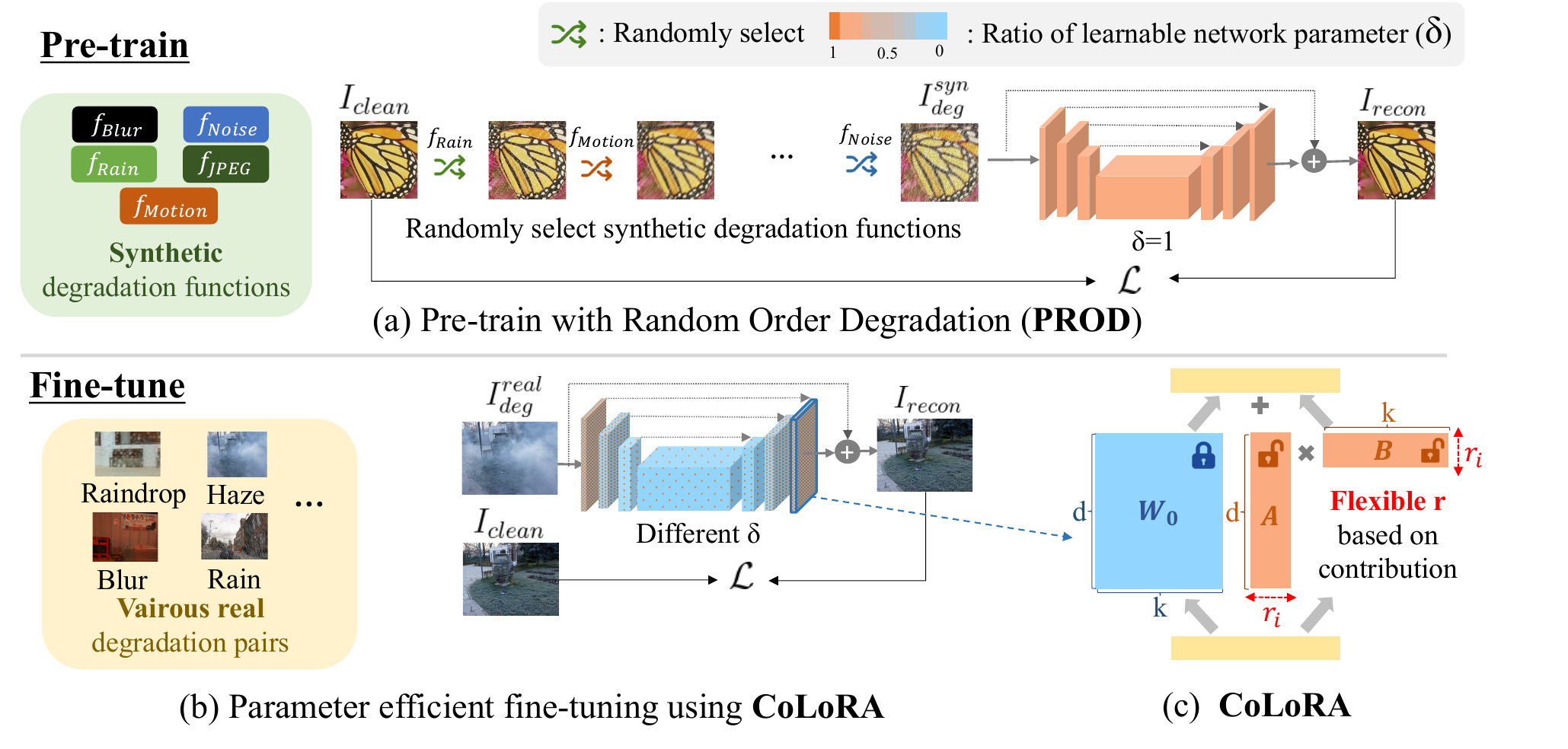}
 %   \vspace{-2.0em}
    \caption{
    The overview of our proposed CoLoRA with PROD. 
    (a) Our PROD leverages high-quality clean images and synthetic degraded low-quality images for pre-training the model. 
    (b) Our proposed Contribution based efficient LoRA (CoLoRA) for new IR tasks.
    The proposed CoLoRA is configured to have different ratio of learnable network parameter ($\delta$) for each layer based on quantified contributions (Sec~\ref{sec:observation}), enabling efficient fine-tuning for new tasks.
    (c) CoLoRA can be adjusted according to contribution.
    }
    \label{Fig2_overview}
\end{figure*}

We propose a Contribution based efficient LoRA (CoLoRA) with Pre-training with Random Order Degradation (PROD) for IR, as illustrated in Fig.~\ref{Fig2_overview}.
%Our overall framework is illustrated in Fig.~\ref{Fig2_overview}. 
Section~\ref{sec:pretrain} introduces the pre-training method PROD, and Section~\ref{sec:observation} investigates the quantified contribution to each layer.
In Section~\ref{sec:CoLoRA}, we propose CoLoRA that adjusts the ratio of learnable network parameter based on contributions.

\subsection{Pre-training with Random Order Degradation (PROD) }
\label{sec:pretrain}
We propose a PROD that randomly applies synthetic degradation to clean images during the pre-training, as illustrated in Fig.~\ref{Fig2_overview}.
Detailed information on distortion functions and magnitudes are in the supplementary materials.
Applying 1 to N synthetic degradations to a clean image, PROD can represent a total of $(H^{N+1} - 1)/(H-1)$ different types of degraded images where N is set to 6 and $H(=5)$ represents the number of degradation functions. 
This PROD method enables about $137 K$ kinds of degradation representations, which is about  $4 K$ times more than the previous single~\cite{chen2021pre,li2021efficient} and fixed order~\cite{liu2023degae} synthetic degradation methods.
Note that a similar random degradation idea was proposed in~\cite{zhang2021designing} and \cite{cubukpractical}, but it was designed for super resolution and classification, not for pre-training for multiple tasks.
The experimental results are in the supplementary material.
The PROD significantly expands the scalability and generalization of the pre-train model, bringing excellent performance in IR with real data.

\begin{figure}[!b]
    \centering
\includegraphics[width=1\textwidth]{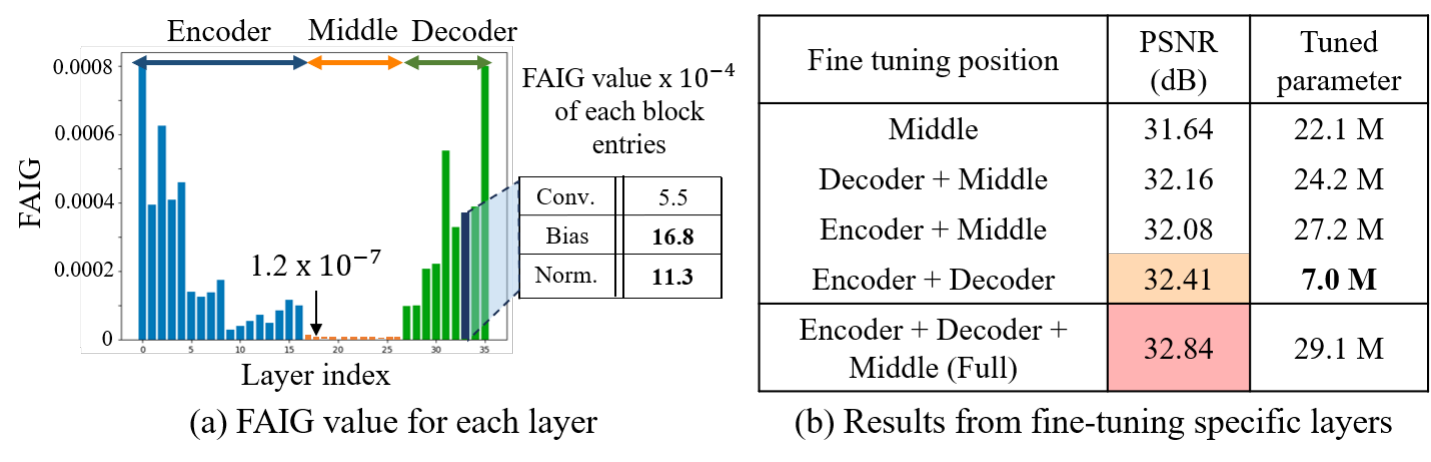}
   % \vspace{-1.em}    
    \caption{(a) For each layer, we measured the FAIG score using a pre-trained model and a fine-tuned model specifically tuned for the specific task to observe its contribution.
    (b) Experimental results according to fine-tuning location for a blur task.
    The encoder and decoder occupy 18\% of the total network parameters, while the middle layers account for 80\%.
    The encoder and decoder have higher FAIG scores than the middle.
    Bias and normalization have higher FAIG values compared to the weight layer (Conv).
    }
    \label{Fig_FAIG}
     %   \vspace{-1.em}
\end{figure}

\subsection{The Key Components of Fine-Tuning for a New Task}
\label{sec:observation}

We utilized Filter Attribution method based on Integral Gradient (FAIG)~\cite{xie2021finding} scores to quantify the major contributing network parts for new IR tasks.
FAIG is measured through Integrated Gradients (IG) calculations using pre-trained and fine-tuned models.
The FAIG calculation for each layer involves the baseline model ($\theta_{ba}$) and the target model ($\theta_{ta}$), defined as follows:
%\vspace{-1em}
\begin{equation}
\begin{split}
\operatorname{FAIG}_i(\theta_{ba}^i, \theta_{ta}^i, x) %\qquad\qquad\qquad\qquad\qquad\qquad 
\approx  \sum_{j=1}\left| \frac{1}{M}[\theta^{i, j}_{ba}-\theta^{i,j}_{ta}] \sum_{t=0}^{M-1}\left[\frac{\partial \mathcal{L}(\rho(\beta_t), x)}{\partial \rho(\beta_t)} \right]_{i,j}  \right|,
\end{split}
\label{eq:faig}
\end{equation}
%\vspace{-0.8em}
\noindent
where $M$ represents the total number of steps in the integral approximation, setting to 100 as in FAIG. $\beta_t$, $j$, $i$ and $\rho$ are $t/M$, the kernel index, layer index and interpolation between models, respectively.
The high value of FAIG indicates a significant importance for new tasks~\cite{park2023all}. 
Conversely, lower values indicate high similarity between the baseline and target model, implying low importance for the new task.
In Fig.~\ref{Fig_FAIG}, we measure the FAIG score of each block using the full fine-tuned  NAFNet~\cite{chen2022simple} on real IR tasks after PROD execution. 
The FAIG values are an average of the 6 real IR tasks.
Fig.~\ref{Fig_FAIG} (a) demonstrates that the encoder and the decoder part of the network have significantly higher FAIG values than the middle layer. 
This indicates that the middle layers, accounting for a substantial fraction (around 80\%) of the total network parameters, have a small contribution compared to the encoder and decoder, which account for a small portion (around 18\%).
Bias and Normalization layers have higher FAIG values than weight layers, despite a smaller portion of the total network parameters.

Based on these observations, we conducted experiments on partially fine-tuning for the new task, as shown in Fig.~\ref{Fig_FAIG} (b).
Fine-tuning only the encoder and decoder demonstrated superior performance with fewer network parameters compared to other methods.
We observed that by fine-tuning the encoder and decoder parts with high FAIG scores, the network performance can be improved without unnecessary tuning to the middle layers.
These experiments suggest that FAIG can be helpful in quantifying how specific parts of the network contribute to new tasks.
These observations lead to our proposed CoLoRA.

\subsection{Contribution-based Low-Rank Adaptation (CoLoRA)}
\label{sec:CoLoRA}

Hu \textit{et al.}~\cite{hu2021lora} proposed a method known as LoRA aimed at fine-tuning only small network parameters. 
The training weight matrix $\triangle W$ is represented as follows:
\begin{equation}
W_0 + \triangle W = W_0 + BA,
\label{eq:faig1}
\end{equation}
where $W_0 \in \mathbb{R}^{d\times k}$, $B \in \mathbb{R}^{d\times r}$, $A \in \mathbb{R}^{r\times k}$, and the $r \ll \min(d,k)$ is pre-trained weight matrix, projection weight matrices, and rank, respectively. In the fine-tuning phase, $A$ and $B$ consist of learnable parameters while $W_0$ remains frozen and does not undergo gradient updates.
The output vectors are coordinate-wise sum of $W_0$ and $\triangle W = BA$, which have the same input.
The conventional LoRA employs a fixed rank value of $r$ for all layers of the network because optimizing the low-rank parameter for each layer is infeasible, thus imposes constraints on the efficient utilization of parameters and limits its overall performance.

To address these limitations, we propose CoLoRA, a method that flexibly adjusts the ratio of learnable network parameter  based on contributions.
In Section~\ref{sec:observation}, we adjust the ratio of learnable parameters $\delta$ by applying different values of $r$ for each layer based on the quantified FAIG score. 
The $\delta$ in CoLoRA, according to the size of $r$, is defined as 
$\delta_i = r_i(d_i+k_i)/(d_i \times k_i)$, where $i$ denotes  the network layer index, and $d$ and $k$ represent the size of $W_0$.
For the FAIG scores measured for each block as in Fig.~\ref{Fig_FAIG} (a), excluding the intro and end layers, we normalize to ensure the maximum value = 1.
Using the normalized FAIG score, we propose a threshold-based scaling to determine $\delta$ without increasing complexity as follows: 
\[
\delta^{s} = 
\begin{cases} 
\text{Norm}(FAIG^{s}) \times \alpha& \text{if } \text{Norm}(FAIG^{s}) > 0.5, \\
\text{Norm}(FAIG^{s}) \times \beta & \text{otherwise},
\end{cases}
\] 
where $\alpha$, $\beta$ and \text{Norm} are two scale factors and a normalization function.
$FAIG^{s}$ denotes the average value of $FAIG_i$ values for each stage and $s$ is the stage index.
We used scale factors ($\alpha,\beta$) to optimize training parameters.
The $\delta$ values are applied uniformly across all tasks.
Thanks to the proposed CoLoRA, only a small fraction of learnable network parameters (approximately 7\%) is allocated for each IR task, resulting in a significant reduction in memory usage compared to full fine-tuning (100\%) for additional tasks.
Our proposed CoLoRA with PROD method yields high performance using less memory compared to the previous work~\cite{liu2023degae} with full-tuning in Section~\ref{sec:Experiments}.
Our proposed CoLoRA can merge with fixed pre-trained model weights with a simple linear model. Therefore, there is no inference delay compared to the full fine-tuning method~\cite{hu2021lora}.
Detailed $\delta$ values for each network are in the supplementary material.

\noindent
\textbf{Bias and Normalization layer.} 
The parameters of Bias and Normalization layers, despite a small portion of the entire network, have high FAIG values. Therefore, we adjust the bias and normalization layer together during fine-tuning. Note that LoRA~\cite{hu2021lora} does not update the Bias and Normalization layers.

\section{Experiments}
\label{sec:Experiments}
\textbf{Experiment setups.} 
We evaluated the proposed  CoLoRA with PROD on 6 \emph{real} IR tasks using CNN-based NAFNet~\cite{chen2022simple} and transformer-based Restormer~\cite{zamir2022restormer}.
To evaluate our method, we compare it with the previously pre-trained methodology, DegAE~\cite{liu2023degae}.
In Restormer, the DegAE~\cite{liu2023degae} uses publicly released pre-trained weights, and in NAFNet, the DegAE~\cite{liu2023degae} was reproduced based on the publicly released code. 
In CoLoRA, we experimentally used $\alpha$ = 1 and $\beta$ = 0.2 for NAFNet, 
and $\alpha$ = 0.75 and $\beta$ = 0.1 for Restormer. 
The learning iteration for the pre-training model and fine-tuning were set to 200K and 10K, respectively.
The learning rate starts at 1e-3 for NAFNet and 3e-4 for Restormer and gradually decreases to 1e-6 according to the cosine annealing schedule.
The evaluation of experiment results was performed using PSNR. 
In pre-training, PSNR loss was used, and in the fine-tuning process, NAFNet and Restormer used PSNR loss and L1, respecitvely.
Detailed information is in the supplementary material.

\noindent
\textbf{Six real-world image restoration task datasets:} 
To evaluate our proposed method, we conducted experiments using 6 real-world Image Restoration (IR) datasets:
 Real Rain, Raindrop, Rain\&Raindrop (RainDS~\cite{quan2021removing}), Noise (SIDD~\cite{abdelhamed2018high}), Haze (SMOKE~\cite{jin2022structure}), and Blur (BSD~\cite{zhong2020efficient}). 
RainDS~\cite{quan2021removing} consists of 120 training data and 100 test data for each task, Rain and Raindrop, Rain and Raindrop. 
SMOKE~\cite{jin2022structure} includes of 120 training images and 12 test images. 
BSD~\cite{zhong2020efficient} comprises  of 18,000 training images and 3,000 test images (2ms–16ms). 
SIDD~\cite{zhong2020efficient} data comprises  of 160 training images and 1,280 validation samples. 
%The validation data consists of a $256 \times 256$ patched images.

\begin{table}[!t]
\caption{Benchmark results for real Rain, Raindrop, Raind\&Raindrop, Haze and Blur test datasets in PSNR (dB).
Our \textit{CoLoRA with PROD} achieved comparable performance on NAFNet and Retsomer, respectively, by updating only a small 2M and 2.9M of tuned network parameters, in contrast to the full fine-tuning of 29M and 26M.
}
 %       \vspace{-0.5em}
\label{benchmark}
\begin{minipage}{1.0\linewidth}
\centering
\begin{tabular}{c|ccc|c}
%\hline
\toprule
Method                    & \ Rain\quad  & \quad Raindrop & \quad Rain\&Raindrop & Tuned par.\\ \hline
SPANet~\cite{wang2019spatial}                     & \ 22.17\quad  & \quad 20.43    & \quad 19.43  & -         \\
PReNet\cite{ren2019progressive}                     & \ 24.56 & \quad 22.33    &\quad  21.20   & -        \\
DRDN~\cite{deng2020detail}                      &\  23.83 &\quad  21.14    &\quad  20.10      & -     \\
CCN~\cite{quan2021removing}                       &\  26.83 &\quad  24.81    &\quad  23.09   & -        \\ \hline
NAFNet (NAF.) + Full~\cite{chen2022simple}                    &\  27.09 &\quad  24.86    &\quad  23.79     & 29M      \\
DegAE (NAF.) + Full~\cite{liu2023degae}            &\  27.22 &\quad  24.96    &\quad  24.09    & 29M       \\
\textbf{PROD (NAF.) + Full}             &\  \colorbox{tabfirst}{27.42} &\quad   \colorbox{tabsecond}{25.02}    &\quad  \colorbox{tabsecond}{24.37}      & 29M     \\
\textbf{CoLoRA w. PROD (NAF.)} &\  \colorbox{tabsecond}{27.37} &\quad  \colorbox{tabfirst}{25.22}    &\quad  \colorbox{tabfirst}{24.62}       & 2M    \\ \hline
Restormer (Rest.) + Full~\cite{zamir2022restormer}                &\ 26.93 &\quad  24.73    &\quad  23.53      & 26M     \\
DegAE (Rest.) + Full~\cite{liu2023degae}             &\ 27.11 &\quad  24.95    &\quad  23.97  & 26M          \\
\textbf{PROD (Rest.) + Full}              &\  \colorbox{tabfirst}{27.48} &\quad  \colorbox{tabfirst}{25.07}    &\quad \colorbox{tabfirst}{24.34}      & 26M     \\
\textbf{CoLoRA w. PROD (Rest.)}  &\  \colorbox{tabsecond}{27.35} &\quad  \colorbox{tabsecond}{24.96}    &\quad \colorbox{tabsecond}{24.04}       & 2.9M    \\ 
\bottomrule
\end{tabular}
 \label{supp:rain}
\end{minipage}
\begin{minipage}{.49\linewidth}
\centering
\begin{tabular}{c|c}
\toprule
Method                    &\ Haze           \\ \hline
DCP~\cite{he2010single}                       &\ 11.25          \\
GDN~\cite{liu2019griddehazenet}                       &\ 15.19          \\
MSBDN~\cite{dong2020multi}                     &\ 13.19          \\
DeHamer~\cite{guo2022image}                   &\ 13.31          \\
SMOKE~\cite{jin2022structure}                     &\ 18.83          \\ \hline
NAF. + Full~\cite{chen2022simple}                    &\ 19.27          \\
DegAE (NAF.) + Full~\cite{liu2023degae}            &\ \colorbox{tabsecond}{20.23}          \\
\textbf{PROD(NAF.) + Full}             &\ \colorbox{tabfirst}{20.33} \\
\textbf{CoLoRA w. PROD(NAF.)} &\ 20.17          \\ \hline
Rest. + Full~\cite{zamir2022restormer}                 &\ 19.67          \\
DegAE (Rest.) + Full~\cite{liu2023degae}             &\ \colorbox{tabfirst}{20.28}          \\
\textbf{PROD (Rest.) + Full}              &\ \colorbox{tabsecond}{20.16} \\
\textbf{CoLoRA w. PROD (Rest.)}  &\ 20.04          \\ \bottomrule
\end{tabular}
\end{minipage}%
\begin{minipage}{.49\linewidth}
\centering
\begin{tabular}{c|c}
\toprule
Method                    &\ Blur           \\ \hline
IFI-RNN~\cite{nah2019recurrent}                       &\ 31.53          \\
ESTRNN~\cite{zhong2020efficient}                       &\ 31.95          \\
CDVD-TSP~\cite{shang2021bringing}                     &\ 32.16          \\
STFAN~\cite{zhou2019spatio}                   &\ 32.19          \\
PVDNet~\cite{son2021recurrent}                     &\ 32.22          \\ \hline
NAF. + Full~\cite{chen2022simple}                    &\ \colorbox{tabsecond}{32.30}          \\
DegAE (NAF.) + Full~\cite{liu2023degae}            &\  32.19     \\
\textbf{PROD (NAF.) + Full}             &\ \colorbox{tabfirst}{32.84} \\
\textbf{CoLoRA w. PROD (NAF.)} &\   32.08    \\ \hline
Rest. + Full~\cite{zamir2022restormer}                 &\      30.42    \\
DegAE (Rest.) + Full~\cite{liu2023degae}             &\  31.47     \\
\textbf{PROD (Rest.) + Full}              &\ \colorbox{tabfirst}{32.38} \\
\textbf{CoLoRA w. PROD (Rest.)}  &\ \colorbox{tabsecond}{31.90}      \\ \bottomrule
\end{tabular}
\end{minipage}
\end{table}
\subsection{Benchmark Results for Real Various IR Tasks}
In Table~\ref{benchmark}, we summarized the benchmark results on the real Rain, Raindrop, Rain\&Raindrop, Haze and Blur datasets to evaluate our proposed \textit{CoLoRA with PROD}.
%To make a fair comparison with previous methods, we trained the model using the entire training dataset from the real Rain, Raindrop, Rain\&Raindrop, Haze and Blur datasets.
To make a fair comparison with previous methods, we trained the model using the entire training dataset from the real various IR task datasets.
For each task, all methods except DegAE and Our \textit{PROD} are initialized randomly without a pre-trained model.
%All methods except \textit{CoLoRA} update the entire network.
All methods except Our \textit{CoLoRA} perform full fine-tuning on new tasks.
In full fine-tuning, NAFNet and Restormer have 29 M and 26 M (100\%) trainable parameters, respectively.
The CoLoRA method have 2M and 2.9 M (7\% and 11\% of the total) trainable parameter in NAFNet and Retormer, respectively.
Our PROD achieved state-of-the-art PSNR on datasets containing Real Rain, Raindrop, Rain\&Raindrop, Haze and Blur.
Our \textit{CoLoRA with PROD} approach achieved similar performance on NAFNet and Retsomer, respectively, by updating only small 2M and 2.9M (7\% and 11\% of the total) network parameters compared to full fine-tuning 29M and 26M (100\%).
These results demonstrate that our proposed \textit{PROD} and \textit{CoLoRA with PROD} methods effectively and efficiently improve various IR performances, with abundant data.

\begin{figure*}[t]
    \centering
    \includegraphics[width=1\textwidth]{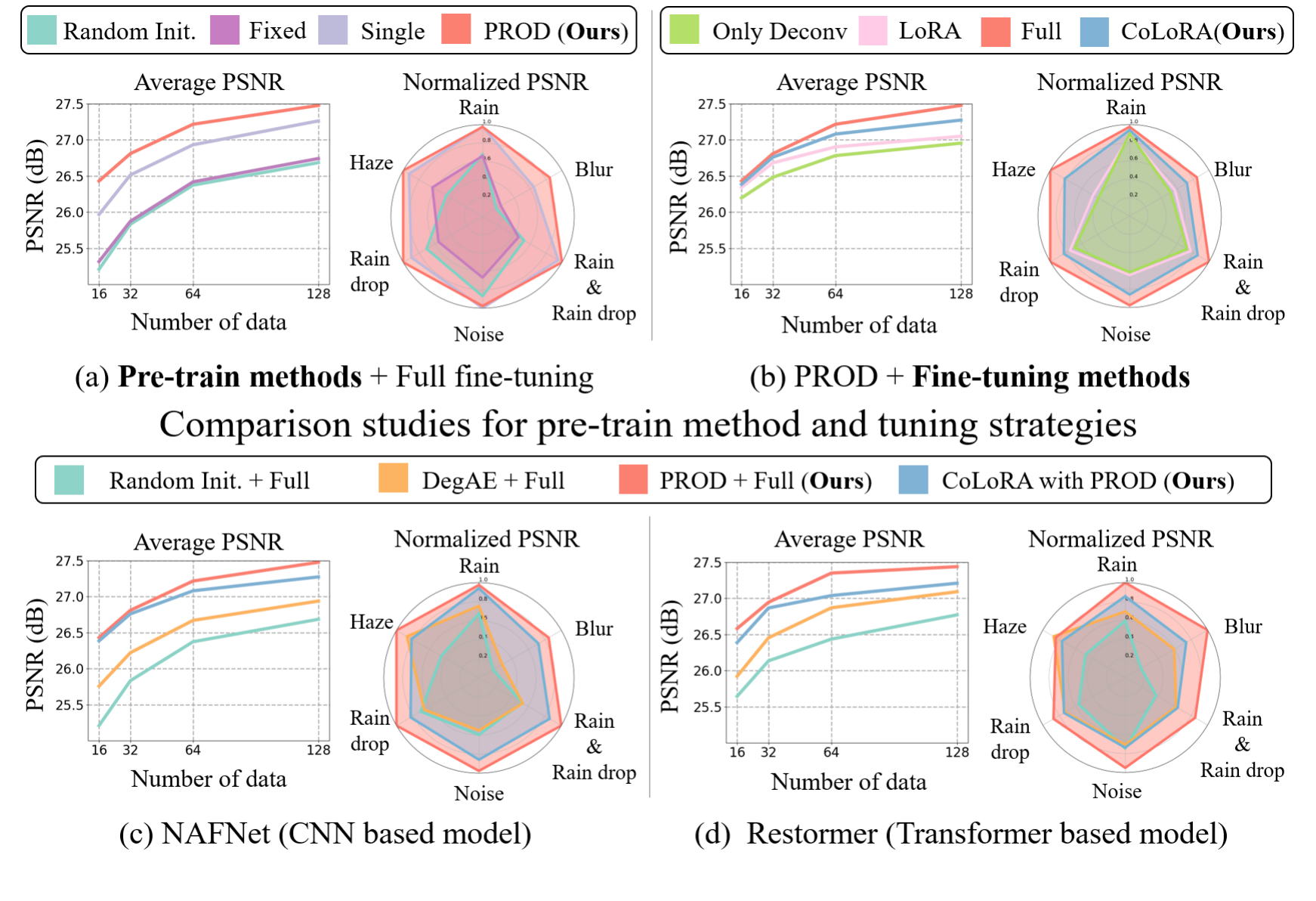}
  %   \vspace{-3em}
    \caption{Performance comparison based on the scale of training data for 6 IR tasks. 
    In the graph, the results of the 6 IR tasks are averaged for comparison.
    The x-axis represents the number of training data, and the y-axis is the average PSNR. 
    In the radar graph, we compare the results of 6 IR tasks with Normalized PSNR at a training data size of 128.  (a) and (b) present experimental results corresponding to pre-training and fine-tuning methods, respectively.  (c) and (d) experimental results for the Our CoLoRA with PROD in NAFNet and Restormer. Our proposed CoLoRA (7\%) has much fewer tuned network parameters compared to the full fine-tuning (100\%) of NAFNet.
}
    \label{abalation}
     %   \vspace{-2em}
\end{figure*}

\subsection{Comparing Proposed Methods in Limited Real-world Scenario}
In Fig.~\ref{abalation}, we conducted a comparative study to demonstrate the performance and efficiency of the parameter-efficient fine-tuning method, CoLoRA, and the proposed pre-training method PROD, depending on the number of training data.
To summarize each experiment, (a) ``Pre-training Methods'' presents the experimental results based on different pre-training methods, (b) ``Tuning Strategies'' presents the results based on efficient parameter tuning methods, and (c)\&(d) ``Different Network Architectures'' provides the experimental results based on pre-training and tuning methods on CNN and vision transformer architectures.
Fig.~\ref{abalation} summarizes PSNR results for six real-world IR tasks with respect to the number of training data during fine-tuning.
The number of training data for each tasks were set to [16, 32, 64, 128]. 
These small datasets reflect the reality of IR tasks: acquiring paired real world datasets is often difficult and expensive (e.g., weather, medical imaging). 
In Fig.~\ref{abalation} (line graph), the y-axis represents the average PSNR results of six IR tasks, while the x-axis represents the number of training data during fine-tuning. 
Each vertical line in the line graph represents results from experiments fine-tuning a total of 6 IR tasks with 4 different amounts of training data.
Fig.~\ref{abalation} (radar graph) represents the normalized PSNR results when number of training data is 128 for six IR tasks with real test data.

\noindent
\textbf{Comparison of pre-training methods.}
To validate the effectiveness of our proposed PROD method, we conduct a comparative study based on pre-training methods, using the CNN-based NAFNet, as illustrated in Fig.~\ref{abalation} (a).
\textit{Random Init} refers to the method of randomly initializing the network parameters.
\textit{Single} denotes methods like IPT~\cite{chen2021pre} and EDT~\cite{li2021efficient}, where various synthetic degradations are randomly applied once. 
\textit{Fixed} represents methods like DegAE~\cite{liu2023degae}, where diverse synthetic degradations are applied in a fixed order. 
\textit{PROD} (Ours) is a method that involves randomly applying various synthetic degradations multiple times to generate low-quality images.
We used full fine-tuning to evaluate the performance of the pre-trained model.
Our \textit{PROD} yielded substantially higher performance than other methods.
Using 32 training data, our \textit{PROD} outperformed \textit{Random Init} and \textit{Fixed}, which utilized 128 training data, by a significant improvement of 0.05 dB and 0.10 dB, respectively.
Our \textit{PROD} method effectively improves performance with limited training data.

%\vspace{0.5em}
\noindent
\textbf{Comparison of fine-tuning strategies.}
In the Fig.~\ref{abalation} (b), 
 we validated our proposed CoLoRA strategy by comparing different tuning methods for a new task using CNN-based NAFNet and the PROD pre-trained weight.
We investigated the following tuning methods : Full fine-tuning (\textit{Full}), Only tunes only the decoder (\textit{Only Decoder}), apply LoRA to all layers (\textit{LoRA}), and our \textit{CoLoRA}.
The \textit{Full} method has 29 M trainable parameters. The \textit{Only Decoder}, \textit{LoRA}, and \textit{CoLoRA}(Ours) have approximately 2 M trainable parameters.
Using 64 training data, our \textit{CoLoRA} outperformed \textit{Only Deconv} and \textit{LoRA}, which utilized 128 training data, by a significant improvement of 0.08 dB and 0.15 dB, respectively.
Our \textit{CoLoRA} method enhances performance efficiently, even with limited training data and a sparse number of learnable parameters.

\begin{figure*}[t]
    \centering
    \includegraphics[width=1.0\textwidth]{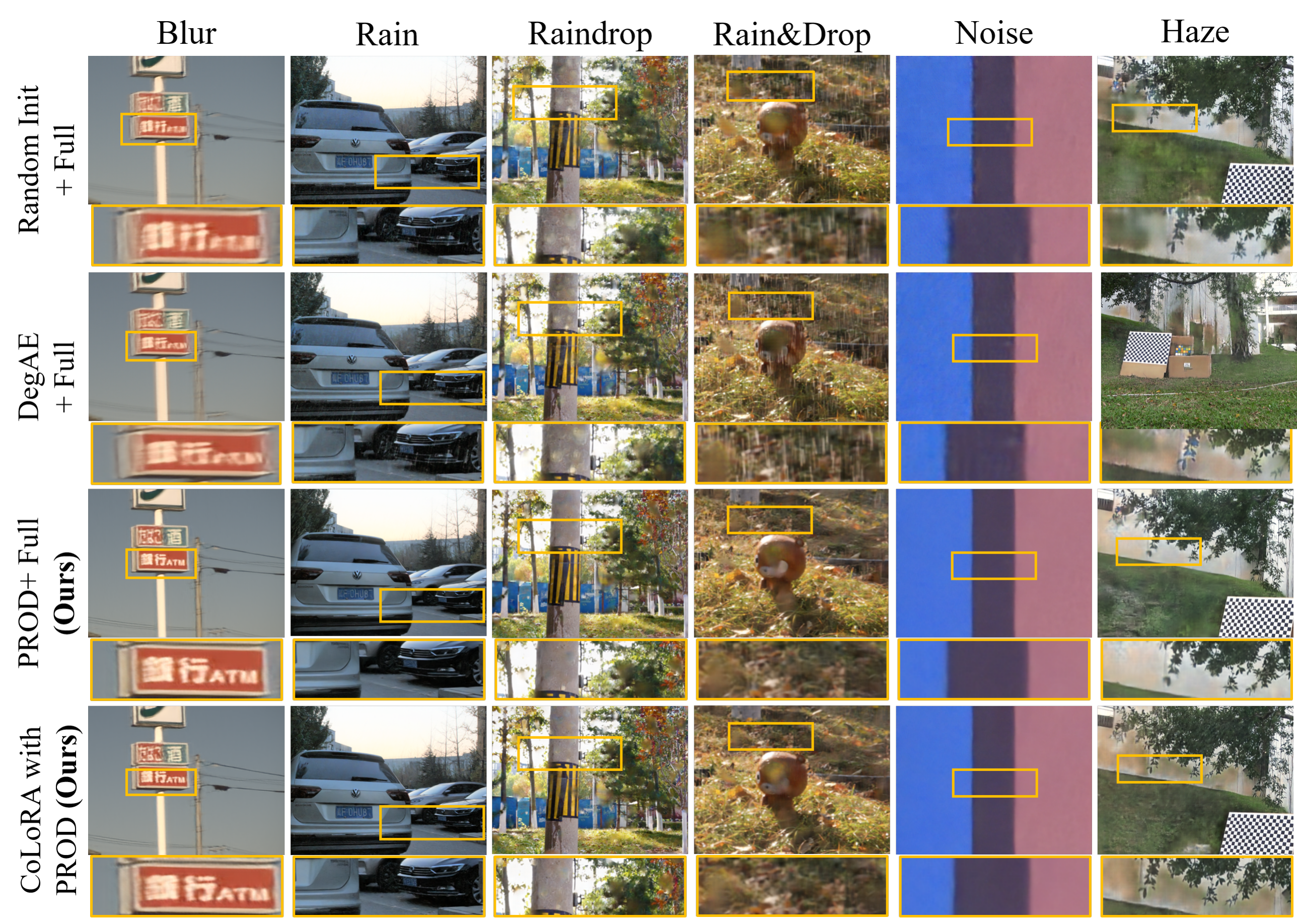}
      %  \vspace{-1em}
    \caption{Qualitative results evaluated on the 6 IR tasks for our proposed method, generic Random initial + Full tuning and DegAE + Full tuning. Our methods with partial and full tuning yielded visually excellent results for the real IR task, outperforming others.}
    \label{final_results}
   % \vspace{-1.0em}
\end{figure*}

%\vspace{0.5em}
\noindent
\textbf{Comparison of different network architectures.}
In Fig.\ref{abalation} (c) and (d), we evaluate the efficiency of the proposed CoLoRA with PROD using the CNN-based NAFNet and Transformer-based Restormer network architectures.
We investigated the following pre-train and tuning methods: Random initialization with full fine-tuning (\textit{Random-init+Full}), DegAE~\cite{liu2023degae} with full fine-tuning (\textit{DegAE+Full}), the proposed full fine-tuning with PROD(\textit{PROD+Full}), and the proposed CoLoRA with PROD method (\textit{CoLoRA with PROD}).
Full fine-tuning has 29M (NAFNet) and 26M (Restormer) trainable parameters, while CoLoRA has 2M (NAFNet) and 2.9M (Restormer).
Our proposed methods (\textit{PROD+Full} and \textit{CoLoRA with PROD}) outperforms than previous methods in 6 IR tasks for both NAFNet and Restormer. 
Using 64 training data, our proposed \textit{CoLoRA with PROD} outperformed \textit{DegAE}~\cite{liu2023degae} and \textit{Random Init} which utilized 128 training data, by a significant improvement of 0.22 dB and 0.36 dB in NAFNet, respectively.
Furthermore, in Restormer, our proposed \textit{CoLoRA with PROD} showed superiority over the \textit{Full with DegAE}~\cite{liu2023degae} in most IR tasks.
Our \textit{CoLoRA with PROD} improves performance efficiently with limited training data and fewer learnable parameters. 
Fig.~\ref{final_results} illustrates the results for 6 IR task with test data according to previous methods and our proposed methods, with the experiments conducted using the NAFNet.
Our methods seems to yield better restoration results than \textit{Full with Random } and \textit{Full with DegAE}~\cite{liu2023degae}.

%\vspace{0.5em}
\noindent
\textbf{Additional experiments on low-light and underwater tasks.}
We have further evaluated on more datasets, underwater (LSUI~\cite{peng2023u} dataset) and low-light enhancement (LoL~\cite{wei2018deep} dataset) tasks. 
All experiments were performed with NAFNet. 
The numerical results for each task are as follows: 
(1) for LSUI data, the PSNR (dB) values of \{NAFNet(Random), DegAE, PROD, CoLoRA with PROD\} are \{22.29, 22.69, 23.32, 22.98\}, respectively.
(2) for LoL data, the PSNR (dB) values of \{NAFNet(Random), DegAE, PROD, CoLoRA with PROD\} are \{22.79, 22.77, 22.96, 22.94\}, respectively.
Our methods achieved high performance, demonstrating the generality.
% 학습 데이터는 총 485개를 사용함.

\subsection{Ablation Study On Scale Values ($\alpha$ and $\beta$) of CoLoRA}
In Table~\ref{scale_results}, we investigated the PSNR results based on CoLoRA's scaling factors  ($\alpha$,$\beta$) for NAFNet and Restormer, to  optimize network parameters.
Selecting $\alpha$ and $\beta$ depends on the trade-off between performance and memory. 
We found that using small values for $\beta$ (0.1-0.2) was efficient while using large values for $\alpha$ (0.7-1.0) was effective.
We selected $\alpha=1, \beta = 0.2$ for NAFNet and $\alpha=0.75, \beta = 0.1$ for Restormer for the highest performance efficiently.
With these criteria, $\alpha$ and $\beta$ can be practically chosen depending on the target device and the applied task.

\begin{table}[!t]
\caption{
Ablation study according to CoLoRA scale parameters $\alpha$ and $\beta$ for Blur task.  
We conducted the experiment using NAFNet and used all training data.
}
\centering
\begin{tabular}{c|cccccc}
\toprule
$\alpha$ / $\beta$           & \quad 1.0 / 1.0   & \quad 1.0 / 0.2 & \quad 1.0 / 0.1  & \quad 0.5 / 0.2   & \quad 0.25 / 0.2  & \quad 0.1 / 0.1   \\\hline
PSNR (dB)       & \quad 32.26 & \quad 32.08 & \quad 31.98&\quad 31.90 & \quad 31.86 & \quad 31.54 \\
Tuned param. & \quad 5.2 M & \quad 1.9 M &\quad 1.6 M &\quad 1.6 M & \quad 1.4 M & \quad 0.8 M \\
\bottomrule
\end{tabular}
\label{scale_results}
   % \vspace{-1.0em}
\end{table}

\section{Discussion}

\begin{figure}[!t]
    \centering
\includegraphics[width=0.9\textwidth]{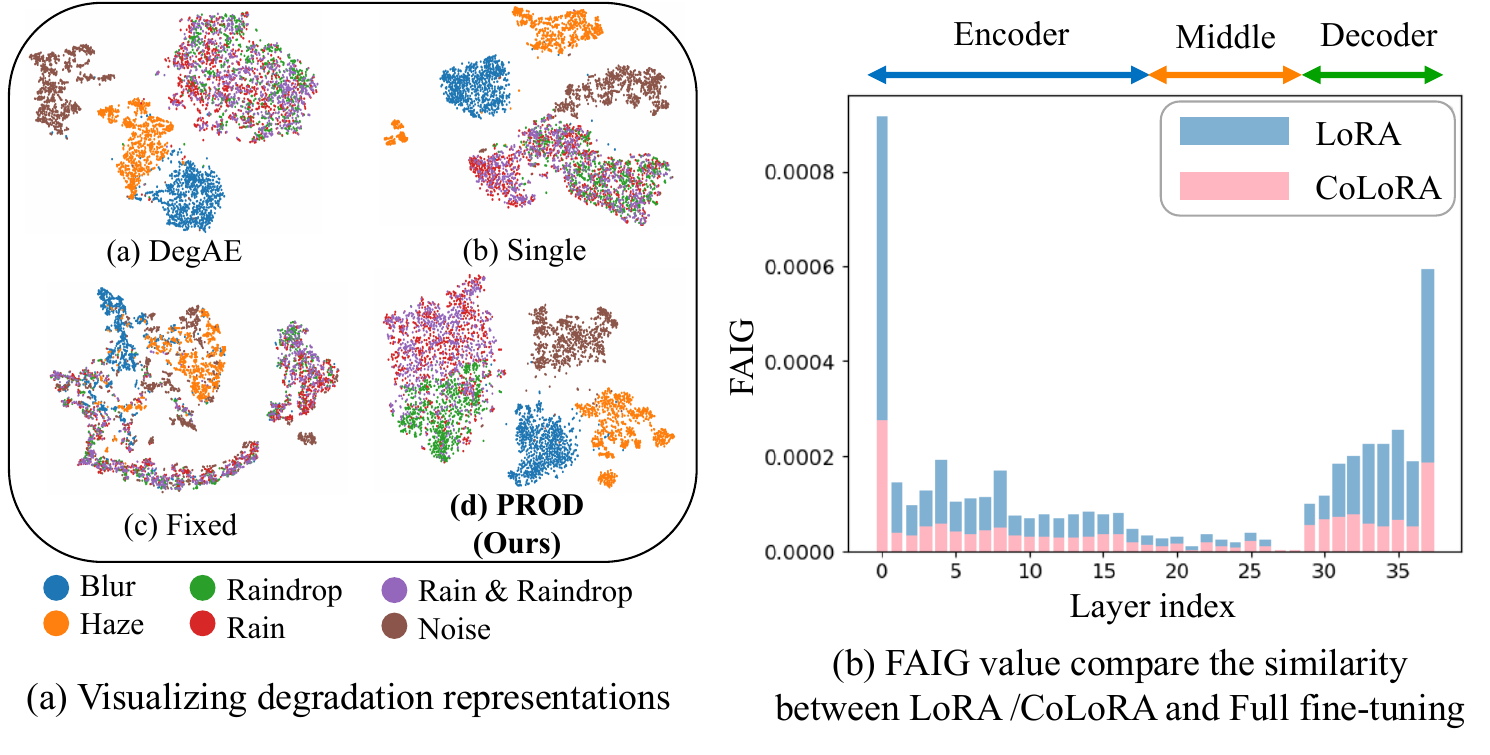}
    %\vspace{-2.0em}
    \caption{(a)  Visualizing degradation representations for 6 real IR tasks using t-SNE without fine-tuning, our PROD method generates more discriminative clusters for untrained data than other pre-training methods.
    (b)  Average FAIG value to compare the similarity between LoRA / CoLoRA and Full-tuning (the lower, the more similar).
}
  %  \vspace{-0.5em}
    \label{TSNE}
   % \vspace{-1.0em}
\end{figure}

%\vspace{0.5em}
\noindent
\textbf{Empirical analysis on pre-train methods without any fine-tuning:}
\label{sec:Empiricalprod}
To observe the scalability and generalization capabilities of our pre-training method, PROD, we conduct an empirical analysis on 6 real IR tasks without any fine-tuning.
The investigated pre-training methods consist of untrained Fixed, Single, DeGAE, and PROD.
Table~\ref{table:PSNR} summarizes the PSNR for 6 real IR tasks based on pre-training methods without any fine-tuning. 
Our proposed PROD achieves better results even when not trained for degradation, surpassing previous approaches.
Fig.~\ref{TSNE} (a) illustrates the degradation representations of  various IR tasks using t-SNE, based on pre-training methods without fine-tuning. 
Features are extracted from the last layer of the middle block of the pre-trained NAFNet for t-SNE plotting, followed by global average pooling, and
t-SNE was plotted for 100 samples.
We believe that the embedded scalability and generalization ability of PROD shows promising performance on the above IR results.

\begin{table}[!t]
\centering
\caption{We evaluate the performance comparison in terms of PSNR (dB) using the pre-trained model without any fine-tuning on 6 real IR tasks.}
   % \vspace{-0.5em}
\begin{tabular}{c|cccccc|c}
\toprule
Method & Rain  & Raindrop & Rain\&Raindrop  & Noise & Blur  & Haze                       & Avg. \\ \hline
Fixed  & 21.76 & 18.81  & 17.42 & 23.68 & 27.49 & \cellcolor{tabfirst}13.06 & 20.37     \\
Single & 22.11 & 18.77  & 17.44 & 23.84 & 27.47 & 13.03 &  20.44    \\ \hline
PROD   & \cellcolor{tabfirst}24.11 & \cellcolor{tabfirst}18.85  & \cellcolor{tabfirst}18.14 & \cellcolor{tabfirst}25.10  & \cellcolor{tabfirst}27.75 & \cellcolor{tabfirst}13.06 &   \cellcolor{tabfirst}21.17  \\ \bottomrule
\end{tabular}
%\end{adjustbox}
\label{table:PSNR}
   % \vspace{-1em}
 \end{table}

%\vspace{0.5em}
\noindent
\textbf{Why CoLoRA, practical deployment:}
Considering that IR can be executed immediately after image acquisition and used for on-device AI, the efficiency, robustness, and scalability of CoLoRA are very important.
%Considering that IR is one of the first tasks right after image acquisition CoLoRA is crucial. 
Compared to full-tuning, CoLoRA excels in efficiency, robustness, and scalability.
For six tasks, CoLoRA has fewer parameters than Full and LoRA (41M for CoLoRA, 41M for LoRA, 174M for Full-tuning), and requires less memory to store (167MB for CoLoRA, 176MB for LoRA, 696MB for Full-tuning), providing significant advantages to on-device AI applications with limited memory~\cite{cai2020tinytl}.
In scenarios with increasing learning rates (2e-4 to 8e-3), CoLoRA demonstrates robust performance over Full-tuning (CoLoRA 25.3 to 22.3dB, Full-tuning 25.3 to 20.8dB).
Additionally, CoLoRA is easier to scale to new tasks compared to all-in-one IR methods.
 CoLoRA can be used in a variety of industries with AI edge devices.

\noindent
\textbf{Empirical Analysis of CoLoRA and LoRA:}
In Fig.~\ref{TSNE}(b), we measured the FAIG value to analyze the correlation between Full Fine-tuning and CoLoRA/LoRA.
Note that lower FAIG score in Eq.(1) indicates more similarity between two networks, reflecting smaller differences and gradients.
CoLoRA has a lower FAIG value compared to LoRA, which explains its higher similarity to full-fine tuning and superior performance.

\noindent
\textbf{Limitation:}
While our CoLoRA has been successfully validated for various downstream IR tasks, our CoLoRA faces the limitation of increasing parameters as the tasks expand even though they are still more efficient than full-tuning.

\section{Conclusion}
We proposed CoLoRA with PROD, pre-training with synthetic data and efficient parameter tuning in both abundant and limited real-data scenarios for image restoration.
Our PROD yielded excellent pre-trained model as compared to prior arts and our CoLoRA enabled efficient fine-tuning only about 7\% learnable parameters.
We demonstrate that our proposed method is flexible enough to work with diverse network architectures and achieves state-of-the-art performance on various IR tasks with real-world datasets.

\section*{Acknowledgements}
This work was supported by the New Faculty Startup Fund from Seoul National University, Institute of Information \& communications Technology Planning \& Evaluation (IITP) grant funded by the Korea government(MSIT) [NO.RS-2021-II211343, Artificial Intelligence Graduate School Program (Seoul National University)], the National Research Foundation of Korea(NRF) grant funded by the Korea government(MSIT) (No. NRF-2022R1A4A1030579), and Creative-Pioneering Researchers Program through Seoul National University. Also, the authors acknowledged the financial support from the BK21 FOUR program of the Education and Research Program for Future ICT Pioneers, Seoul National University.

\bibliographystyle{splncs04}
\bibliography{main}

\clearpage

\appendix

%%%%%%%%% TITLE

\setcounter{equation}{0}
\setcounter{figure}{0}
\setcounter{table}{0}
\setcounter{page}{1}
\makeatletter
\renewcommand{\theequation}{S\arabic{equation}}
\renewcommand{\thefigure}{S\arabic{figure}}
\renewcommand{\thetable}{S\arabic{table}}
\renewcommand\thesection{S\arabic{section}} % S added to 

%%%%%%%%%%%
\begin{center}
\bigskip 
\bigskip 
\textbf{\Large Supplementary Material \\}
\bigskip 
\bigskip 
 
\end{center}%

\section{Details on the proposed CoLoRA with PROD}

\begin{figure}[!h]
    \centering
%              \vspace{-1em}
\includegraphics[width=1\textwidth]{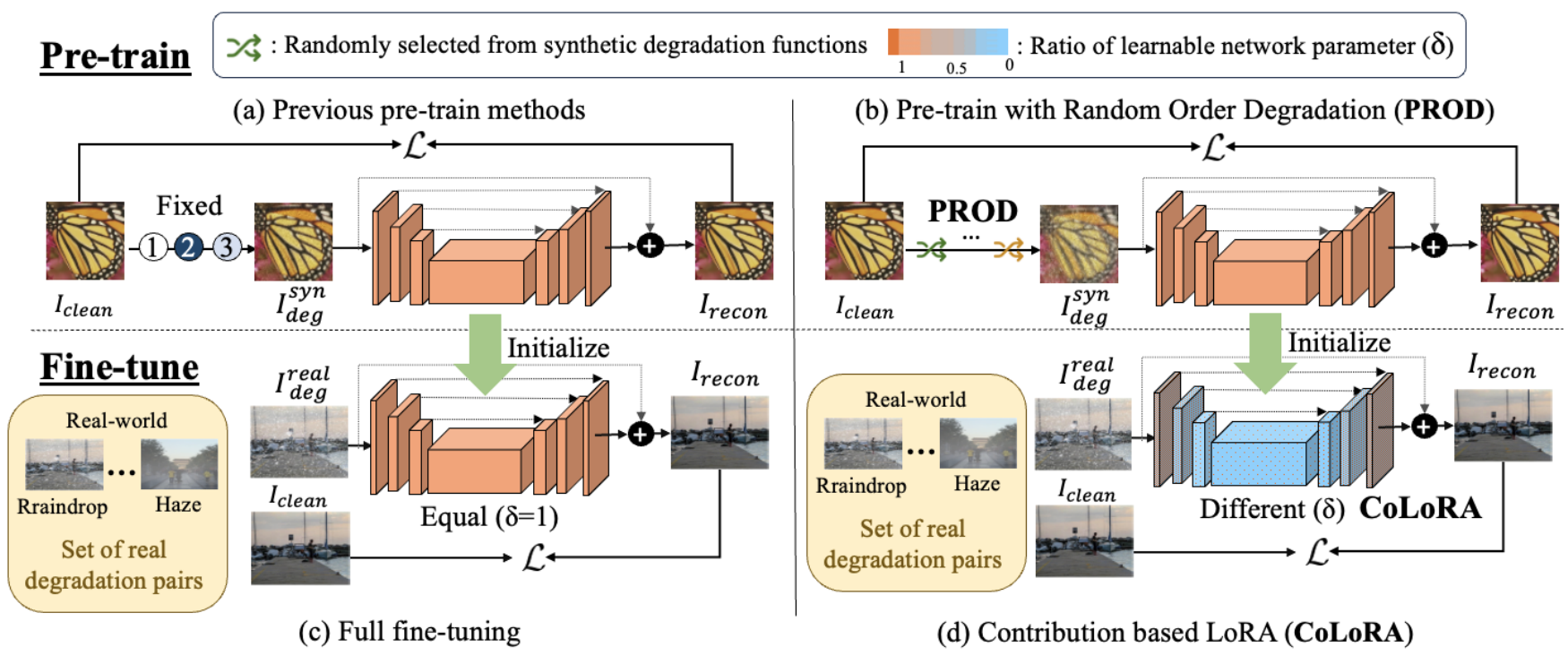}
 %   \vspace{-1.0em}
    \caption{Illustrations of pre-train methods and tuning strategies for image restoration. 
    From synthetic degradation functions, (a) while prior works either randomly select a single degradation~\cite{chen2021pre,li2021efficient} or pre-determine multiple degradations in a fixed order~\cite{liu2023degae}, (b) our proposed PROD pre-trains with generated low-quality images by random order degradations. 
    (c) While utilizing the full fine-tuning strategy is common among prior works, (d) our CoLoRA enables parameter-efficient fine-tuning by freezing the pre-trained model and only adjusting the additional adapter for each task.}
    \label{supp:main}
%        \vspace{-1em}
%        \setlength{\intextsep}{5pt} % 원하는 간격을 여기에 입력하세요
\end{figure}

In Figure~\ref{supp:main}, we compare our CoLoRA - PROD method with existing methods~\cite{chen2021pre,li2021efficient,liu2023degae}.  
Figure~\ref{supp:main} illustrates the differences between PROD, which introduces degradation during pre-training, and other prior works, as well as the distinctions between CoLoRA, a efficient tuning approach, and simple full fine-tuning. 
 Specifically, in simple full fine-tuning, the entire network is tuned. 
 In contrast, our CoLoRA approach is a parameter-efficient tuning methodology that flexibly adjusts the ratio of learnable network parameters based on layer positions.
While prior works must have a separate full fine-tuned network for each novel IR task, our  proposed CoLoRA only need to store the small tunable adaptor for each task (approximately 7\% of the entire network parameters for comparable performance to full-size tuning), so that it will be advantageous in terms of memory for multiple target tasks.

\begin{figure}[!b]
 %   \vspace{-1.5em}    
    \centering
    \includegraphics[width=0.70\textwidth]{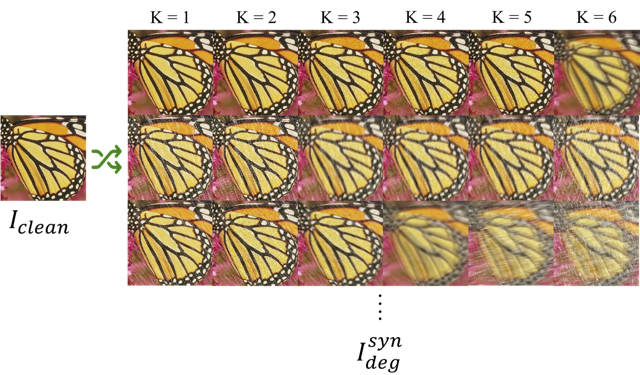}
%     \vspace{0.5em}
    \caption{Example of degraded image generation using PROD with synthetic degradation functions such as Gaussian noise, Gaussian blur, JPEG artifact, and so on.}
    \label{sup:synthetic}
  %      \vspace{-0.8em}
\end{figure}

\section{Details on synthetic degradation functions of PROD}
The environmental settings for synthetic blur, noise and JPEG functions were similar to DegAE~\cite{liu2023degae}.
In the synthetic blur function ($f_{Blur}$), we employ  Gaussian kernels, generalized Gaussian kernels, and plateau-shaped kernels. 
The kernel size is randomly chosen from the set {7, 9, ..., 21}.
For the generalized Gaussian and plateau-shaped kernels, the shape parameters are sampled from the intervals [0.5, 4] and [1, 2], respectively. 
In the synthetic noise function ($f_{Noise}$), we use Gaussian and Poisson, the range for the noise sigma is set to [1, 30], and the Poisson noise scale is set to [0.05, 3].
In the synthetic JPEG function ($f_{JPEG}$), the quality factor is specified within the range [30, 95].
In the synthetic motion blur function ($f_{Motion}$), we use the centered motion kernel proposed by Rim~\cite{rim2020real}. The kernel size is randomly chosen from the set {5, 7, 9, ..., 31}.
In the synthetic rain function ($f_{Rain}$), the rain noise is set to a range of [10, 1000], the rain length is set to [10, 90], the alpha value is chosen between [0.3, 1.3], and the rain angle is set to [-80, 80].
Figure~\ref{sup:synthetic} illustrates synthetic degraded images ($I_{deg}^{syn}$) using PROD.

\section{Ablation studies for CoLoRA}
In Table.~\ref{Norm}, we performed an additional ablation study on bias and normalization for deblurring.
``Bias \& Norm'' improves performance by 0.08 dB with additionally tuned parameter (0.1M), which is consistent with our FAIG analysis. The level of importance could be different for various tasks. 

\begin{table}[!h]
\caption{
Ablation study according to “Bias and Norm.” (B\&N) and “Adaptation”(A.) of CoLoRA using NAFNet in Blur task.}
\centering
\begin{tabular}{|c|cc|c|}
\hline
Method & w/o (Adaptation) & w/o (Bias\&Norm) & CoLoRA \\ \hline
PSNR   & 30.88 dB        & 32.00 dB           & 32.08 dB  \\ \hline
Tund. Param.  & 0.1M        & 1.8M           & 1.9M  \\ \hline
\end{tabular}
\label{Norm}
\end{table}

\section{Details on the $\delta$ of the proposed CoLoRA.}
\begin{figure}[!b]
    \centering
    \includegraphics[width=0.7\textwidth]{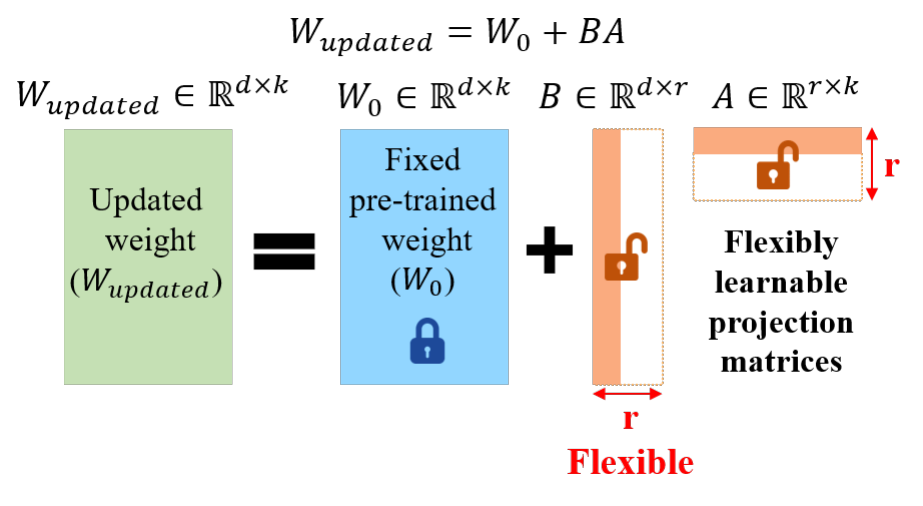}
    \caption{Illustration of our proposed reparametrization method, CoLoRA. The existing LoRA method uses a fixed value of $r$ for all layers since it is challenging to optimize each value at each layer. On the other hand, our proposed CoLoRA adjusts the ratio of learnable parameters ($\delta$) by using different values of $r$ layer by layer without much increasing the degree of freedom.}
    \label{sup:CoLoRA_rep}
 %       \vspace{-1em}
\end{figure}

The LoRA~\cite{hu2021lora} method uses a fixed value of $r$ for all layers.
On the other hand, our proposed CoLoRA adjusts the ratio of learnable parameters ($\delta$) by using  different values of $r$ depending on the layer location, as shown in Figure~\ref{sup:CoLoRA_rep}.
 The $\delta$ in CoLoRA, according to the size of $r$, is defined as follows:
 $\delta_i = (r_i(d_i+k_i))/(d_i \times k_i)$, where $i$ denotes  the network layer index, and $d$ and $k$ represent the size of $W_0$.
The proposed method has designed the size of $r$ to use a small ratio of learnable network parameters (approximately 7\%) in the entire network for each task. 
Thanks to CoLoRA, additional IR operations use only small network parameters, significantly reducing memory usage.
Table~\ref{sup:FAIG} shows Norm(FAIG) values according to location in NAFNet and Restormer.
For real 6 IR tasks, Norm(FAIG) values show the average value and standard deviation.
The middle part has a very small value compared to the encoder part and decoder part.

\begin{table}[t]
\centering
\caption{
In Real 6 IR tasks, these are the Norm(FAIG) average and standard deviation values according to location in NAFNet and Restormer networks.
}
\begin{tabular}{c|ccccccccc}
\toprule
Network                   & \multicolumn{9}{c}{NAFNet}                                                                                                                                                                                                     \\ \hline
\multirow{2}{*}{Location} & \multicolumn{4}{c|}{Encoder}                                                            & \multicolumn{1}{c|}{\multirow{2}{*}{Middle}} & \multicolumn{4}{c}{Decoder}                                                           \\
                          & 1     & 2   & 3                          & \multicolumn{1}{c|}{4}                       & \multicolumn{1}{c|}{}                        & 1     & 2                          & 3                       & 4                      \\ \hline 
Norm(FAIG)                & 0.794 & 1.0 & 0.563                      & \multicolumn{1}{c|}{0.352}                   & \multicolumn{1}{c|}{0.089}                   & 0.391 & 0.831                      & 0.932                   & 0.920                  \\ 
Standard deviation                & 0.229 & 0.241 & 0.172                      & \multicolumn{1}{c|}{0.144}                   & \multicolumn{1}{c|}{0.037}                   & 0.153 & 0.158                      & 0.108                   & 0.201                  \\ \bottomrule \bottomrule
Network                   & \multicolumn{9}{c}{Restormer}                                                                                                                                                                                                  \\ \hline
\multirow{2}{*}{Location} & \multicolumn{3}{c|}{Encoder}             & \multicolumn{1}{c|}{\multirow{2}{*}{Middle}} & \multicolumn{3}{c|}{Decoder}                                                      & \multicolumn{2}{c}{\multirow{2}{*}{Refinement}} \\
                          & 1     & 2   & \multicolumn{1}{c|}{3}     & \multicolumn{1}{c|}{}                        & 1                                            & 2     & \multicolumn{1}{c|}{3}     & \multicolumn{2}{c}{}                            \\ \hline
Norm(FAIG)                & 0.968 & 1.0 & \multicolumn{1}{c|}{0.415} & \multicolumn{1}{c|}{0.005}                   & 0.240                                        & 0.723 & \multicolumn{1}{c|}{0.850} & \multicolumn{2}{c}{0.901}                       \\
Standard deviation                & 0.118 &  0.119 & \multicolumn{1}{c|}{0.131} & \multicolumn{1}{c|}{0.001}                   & 0.095                                        & 0.104 & \multicolumn{1}{c|}{0.041} & \multicolumn{2}{c}{0.150}                       \\
\bottomrule
\end{tabular}
    \label{sup:FAIG}
\end{table}

\section{Details of experiment setups}
For a fair comparison, we evaluate the proposed method under the same conditions using PyTorch~\cite{paszke2019pytorch} on an NVIDIA A100 GPU. 
All our experiments were based on the code published by NAFNet~\cite{chen2022simple} and Restormer~\cite{zamir2022restormer}. 
In all our experiments, the channel dimension of NAFNet~\cite{chen2022simple} is set to 32, the channel dimension of Restormer~\cite{zamir2022restormer} is configured to 48. 
In NAFNet, the Encoder block is configured as [2,2,4,8], the Middle block has 12 layers, and the Decoder block is structured as [2,2,2,2]. 
In Restormer, the Encoder block is composed of [4,6,6], with 8 Middle blocks, and the Decoder block is structured as [6,6,4].
We trained the model with the AdamW~\cite{loshchilov2017decoupled} optimizer ($\beta1$ = 0.9, $\beta2$ = 0.9, weight decay 1e-3), and the training patch size is set to $256 \times 256$.
In NAFNet, the batch size is set to 32, and in Restormer, the batch size is set to 4.
The initial learning rate starts at $3e-4$ and gradually decreases to $1e-6$ according to the cosine annealing schedule. 
Data augmentation for training such as random crop, horizontal flip, and 90-degree rotation were used.
In the case of DegAE~\cite{liu2023degae}, we used the publicly weight files for Restormer, and for NAFNet, we implemented the method based on the publicly code.

\textbf{Pre-training:} 
In the pre-training, we generate degraded low-quality image using synthetic degradation functions, a combination of two open datasets, DIV2K~\cite{agustsson2017ntire} and Flickr2K~\cite{Lim_2017_CVPR_Workshops}.

The number of training iterations for all pre-train models is set to 200K.
To train the pre-train model, PSNR loss was used except for the DegAE~\cite{liu2023degae}.

\textbf{Fine-tuning:} 
In Section 4.1 of the paper, fine-tuning is conducted using 10K iterations.
In Section 4.2 of the paper, Fine-tuning is  performed with a different number of training data, and the number of training data are composed of [16, 32, 64, 128], 100K training iterations are used, and all publicly available training data are used.
%Section 4.2 of the paper uses 100K training iterations and performs fine-tuning using all publicly available training data.
In fine-tuning, NAFNet uses PSNR Loss, and Restormer uses L1 loss.
Due to variations in experimental setups, slight gap in results between the original paper may arise, but this does not impact the validation of our method.
We perform experiments in the same environment, including learning rate, decay method, data augmentation and so on, for all image restoration tasks to ensure consistency across all experiments.

\section{Details of evaluation metrics}
To ensure a fair comparison of our proposed method, we evaluated its performance using  PSNR and SSIM evaluation metrics.
Since evaluation metrics on the Y component in the YUV domain or in the RGB domain tends to exhibit similar trends, we exclusively perform certain measurements, such as Rain, Raindrop, and Rain\&Raindrop, on the Y component.
Specifically in the case of benchmark, Rain, Raindrop, and Rain\&Raindrop, we applied evaluation metrics only to the Y component in the YUV domain.
In the benchmark results presented in Table 1 of the paper, to ensure a fair comparison with previous existing methods~\cite{wang2019spatial, ren2019progressive,deng2020detail,quan2021removing}, we applied evaluation metrics to the Y component in the YUV domain for Rain, Raindrop, and Rain + Raindrop.
Conversely, for Blur, Noise, and Haze, we applied evaluation metrics in the RGB domain.
For the ablation study, we measured evaluation metrics in the RGB domain. 
The ablation study results, presented in Figure 4, Figure S.4, Figure S.5, Figure S.6, and Table S.1 within the paper, entailed applying evaluation metrics in the RGB domain for 6 IR tasks.

\begin{table*}[!t]
\begin{adjustbox}{width=0.9999\linewidth}
\begin{tabular}{cccccccccc}
\hline
\multicolumn{2}{c|}{Number of training data}       & \multicolumn{6}{c|}{2}                                                                                                & \multicolumn{1}{c|}{}                       &                                                                             \\ \cline{1-8}
\multicolumn{1}{c|}{Pre-training} & \multicolumn{1}{c|}{Tunning}                & Rain                 & Raindrop             & Rain\&Raindrop     & Blur                 & Noise                & \multicolumn{1}{c|}{Haze}  & \multicolumn{1}{c|}{\multirow{-2}{*}{Avg.}} & \multirow{-2}{*}{\begin{tabular}[c]{@{}c@{}}Tuned\\ Parm.\end{tabular}} \\ \hline \hline 
\multicolumn{1}{c|}{Random}  & \multicolumn{1}{c|}{}                       & 20.76                & 18.33                & 17.05                & 27.58                & 33.35                & \multicolumn{1}{c|}{13.31} & \multicolumn{1}{c|}{21.73}                  &                                                                             \\
\multicolumn{1}{c|}{DegAE~\cite{liu2023degae}}   & \multicolumn{1}{c|}{}                       & 20.62                &\textbf{19.40}                & 18.90                & 28.02                & 34.72               & \multicolumn{1}{c|}{15.52} & \multicolumn{1}{c|}{22.86}                  &                                                                             \\
\multicolumn{1}{c|}{\textbf{PROD}}    & \multicolumn{1}{c|}{\multirow{-3}{*}{Full}} & \textbf{21.09}                & 19.14                & \textbf{18.97}                & 28.67                & \textbf{35.42}                & \multicolumn{1}{c|}{\textbf{16.18}} & \multicolumn{1}{c|}{\textbf{23.25}}                  & \multirow{-3}{*}{174 M}                                                       \\ \hline
\multicolumn{1}{c|}{\textbf{PROD}}    & \multicolumn{1}{c|}{\textbf{CoLoRA}}                 & 21.77                & 18.78                & 18.70                & \textbf{29.26}                & 34.39                & \multicolumn{1}{c|}{15.10} & \multicolumn{1}{c|}{23.00}                  & \textbf{41 M}                                                                      \\  \hline  
                             &                                             &                      &                      &                      &                      &                      &                            &                                             &                                                                             \\ \hline
\multicolumn{2}{c|}{Number of training data}       & \multicolumn{6}{c|}{4}                                                                                                & \multicolumn{1}{c|}{}                       &                                                                             \\ \cline{1-8}
\multicolumn{1}{c|}{Pre-training} & \multicolumn{1}{c|}{Tunning}                & Rain                 & Raindrop             & Rain\&Raindrop     & Blur                 & Noise                & \multicolumn{1}{c|}{Haze}  & \multicolumn{1}{c|}{\multirow{-2}{*}{Avg.}} & \multirow{-2}{*}{\begin{tabular}[c]{@{}c@{}}Tuned\\ Parm.\end{tabular}} \\ \hline \hline 
\multicolumn{1}{c|}{Random}  & \multicolumn{1}{c|}{}                       & 21.70                & 19.23                & 18.52               & 27.55                & 35.62              & \multicolumn{1}{c|}{17.34} & \multicolumn{1}{c|}{23.33}                  &                                                                             \\
\multicolumn{1}{c|}{DegAE~\cite{liu2023degae}}   & \multicolumn{1}{c|}{}                       & 22.72                & 20.59              & 19.56               & 28.22                & 35.97                & \multicolumn{1}{c|}{\textbf{17.74}} & \multicolumn{1}{c|}{24.13}                  &                                                                             \\
\multicolumn{1}{c|}{\textbf{PROD}}    & \multicolumn{1}{c|}{\multirow{-3}{*}{Full}} & \textbf{23.30}                & \textbf{20.94}                & 20.61                & \textbf{29.78}                & 36.62                & \multicolumn{1}{c|}{17.43} & \multicolumn{1}{c|}{\textbf{24.78}}                  & \multirow{-3}{*}{174 M}                                                       \\ \hline
\multicolumn{1}{c|}{\textbf{PROD}}    & \multicolumn{1}{c|}{\textbf{CoLoRA}}                 & 23.25                & 20.84                & \textbf{20.66}                & 29.70                & \textbf{36.71}               & \multicolumn{1}{c|}{17.32} & \multicolumn{1}{c|}{24.74}                  & \textbf{41 M}                                                             \\\hline            
\multicolumn{1}{l}{}         & \multicolumn{1}{l}{}                        & \multicolumn{1}{l}{} & \multicolumn{1}{l}{} & \multicolumn{1}{l}{} & \multicolumn{1}{l}{} & \multicolumn{1}{l}{} & \multicolumn{1}{l}{}       & \multicolumn{1}{l}{}                        &                                                                             \\ \hline
\multicolumn{2}{c|}{Number of training data}       & \multicolumn{6}{c|}{8}                                                                                                & \multicolumn{1}{c|}{}                       &                                                                             \\ \cline{1-8}
\multicolumn{1}{c|}{Pre-training} & \multicolumn{1}{c|}{Tunning}                & Rain                 & Raindrop             & Rain\&Raindrop     & Blur                 & Noise                & \multicolumn{1}{c|}{Haze}  & \multicolumn{1}{c|}{\multirow{-2}{*}{Avg.}} & \multirow{-2}{*}{\begin{tabular}[c]{@{}c@{}}Tuned\\ Parm.\end{tabular}} \\ \hline \hline 
\multicolumn{1}{c|}{Random}  & \multicolumn{1}{c|}{}                       & 22.97                & 20.99                &  19.73                & 27.72                & 36.96                & \multicolumn{1}{c|}{17.40} & \multicolumn{1}{c|}{24.30}                  &                                                                             \\
\multicolumn{1}{c|}{DegAE~\cite{liu2023degae}}   & \multicolumn{1}{c|}{}                       & 23.50                & 21.46                & 19.83                 & 28.58                & 36.89                & \multicolumn{1}{c|}{\textbf{18.36}} & \multicolumn{1}{c|}{24.77}                  &                                                                             \\
\multicolumn{1}{c|}{\textbf{PROD}}    & \multicolumn{1}{c|}{\multirow{-3}{*}{Full}} & 24.21                & \textbf{21.97}                & \textbf{21.37}                 & 30.24                & \textbf{37.72}                & \multicolumn{1}{c|}{18.31} & \multicolumn{1}{c|}{\textbf{25.64}}                  & \multirow{-3}{*}{174 M}                                                       \\ \hline
\textbf{PROD}                         & \multicolumn{1}{|c|}{\textbf{CoLoRA}}                 & \textbf{24.34}                & 21.91                & 21.20                & \textbf{30.26}                & 37.70                & \multicolumn{1}{c|}{18.15} & \multicolumn{1}{c|}{25.49}                  & \textbf{41 M}          \\\hline                                                    
\end{tabular}
\end{adjustbox}
%\vspace{1.0em}
\caption{We evaluated the performance comparison based on pre-training methods and tuning strategies for 6 IR tasks with \emph{real} validation data in terms of PSNR (dB) in RGB domain and tuned network parameter size (Million). ``Random'' and ``Full'' denote randomly initializing the network parameters without a pre-trained model and full fine-tuning, respectively. Our proposed CoLoRA requires a significantly smaller number of network parameters compared to the full tuning method, as it only tunes 7\% of the network parameters.}
\label{table:limit}
\end{table*}
\begin{table}[!ht]
\caption{Ablation study according to CoLoRA scale parameters $\alpha$ and $\beta$ for Blur task. We conducted the experiment using Restormer and used all training data.}
\centering
\begin{tabular}{c|c|ccccccc}
\toprule
$\alpha$ & \multirow{2}{*}{Full} & 1     & 1.0   & 0.75  & 1.0   & \textbf{0.75}  & 0.5   & 0.25  \\
$\beta$  &                       & 1     & 0.2   & 0.2   & 0.1   & \textbf{0.1}   & 0.1   & 0.1   \\ \hline
PSNR (dB)  & 32.38                 & 32.14 & 32.01 & 31.88 & 31.94 & \textbf{31.90} & 31.80 & 31.82 \\
Tuned parameter & 26.2                  & 8.7   & 4.0   & 3.6   & 3.5   & \textbf{2.9}   & 2.5   & 1.9  \\
\bottomrule
\end{tabular}
\label{supp:scale}
\end{table}

\section{Comparison of pre-train methods and tuning strategies for extremely limited  training data}
In this section, we evaluated the proposed CoLoRA with PROD on 6 IR datasets extremely limited  training data and a training iteration is 10K.
%Table~\ref{table:limit} summarizes the performance in PSNR on each task and the number of total parameters.
Table~\ref{table:limit} summarizes the PSNR results for six real IR tasks with extremely small number of training data during fine-tuning in NAFNet.
The extremely small number of training data for each tasks were set to [2, 4, 8]. 
The full fine-tuning method (\textit{Full}) has 29 M trainable parameters in NAFNet~\cite{chen2022simple}.
In NAFNet, the full fine-tuning (\textit{Full}) method requires 6 times more network parameters $174 M$ ($29 M \times 6$)  than the original network parameters $29 M$ due to explicitly separated structures for multiple degradations, and our proposed CoLoRA uses an additional $2 M$ parameters for a single task, resulting in a total of $41 M$ ($29 M + 2M \times 6$) parameter for the network required in the 6 IR tasks.
These results demonstrate that the our proposed \textit{PROD + Full} and \textit{PROD + CoLoRA} method effectively and efficiently improves various IR performances, with extremely limited data.

\subsection{Ablation study on scales ($\alpha$ and $\beta$) of CoLoRA (Restormer)}
In Table~\ref{supp:scale}, we investigated the PSNR results based on the scaling factors ($\alpha$,$\beta$) of CoLoRA to further optimize the tuned network parameters.
All experiments are conducted using the transformer-based Restormer with full training data.
As the values of $\alpha$ and $\beta$ increase, the performance improves, and accordingly, the tuned network parameters also increase.
We have experimentally selected ($\alpha=0.75, \beta = 0.1$) that efficient and brings the highest performance.
Through scale adjustment, it is possible to use the optimal parameters for each task.

\section{Details on tuning strategies in NAFNet}
Table~\ref{supp:parameter} provides detailed tuned parameters for experiments based on tuning strategies. 
The tuned parameter are in millions (M). 
The \textit{Full} method has 29 M trainable parameters. The \textit{Only Decoder}, \textit{LoRA}, and \textit{CoLoRA}(Ours) have approximately 2 M trainable parameters.
The LoRA~\cite{hu2021lora} is constructed based on open-source code, with the value of $r$ set to 16 for all layers.

\begin{table}[!h]
\caption{
In Section 3.1, ``tuning parameters'' of this paper refers to the number of tuned parameters in millions (M). }
\centering
\begin{tabular}{c|c}
\toprule
Tuning strategies      & Tuned parmeter (M) \\ \hline
Only Deconv &   1.942    \\
LoRA~\cite{hu2021lora}        &   2.442    \\
\textbf{CoLoRA (Ours)}     &  1.993        \\
Full        &     \textbf{29.160} \\
\bottomrule
\end{tabular}
%\vspace{1.0em}
\label{supp:parameter}
\end{table}

\section{Our proposed PROD vs BSRGAN }
Indeed, BSRGAN developed another pre-trained method with multiple degradations only for SR, while our PROD is a method for multiple IR tasks. Nevertheless, we performed experiments using 64 fine-tuning datasets on rain/drop and blur,
demonstrating the superiority of our work over BSRGAN (PROD 31.4dB vs BSRGAN 30.8dB for deblur and PROD 22.7dB vs BSRGAN 22.1dB for derain).

\section{Benchmark results}
We evaluated our proposed CoLoRA with PROD on the mixed Snow\&Haze~\cite{chen2021all} benchmark dataset in Table~\ref{supp:snow}. 
The mixed Snow\&Haze~\cite{chen2021all} benchmark dataset contains 10,000 training images and 3,000 test images.  
"All-in-One image restoration" may experience a decrease in performance as the number of tasks increases, and there are inherent limitations in extending it to new tasks.
On the other hand, "Efficient Fine-tuning" maintains consistent results even as the number of tasks increases, and it is easily scalable to accommodate new tasks.
In Park~\cite{park2023all}, the results are presented for All-in-One image restoration methods when handling only three tasks.
Thanks to CoLoRA with PROD, the proposed method in NAFNet architecture provides the highest PSNR compared to existing methods, even though only a small ratio of tuned parameters are learned in the entire network.

\begin{table}[h]
\centering
\caption{Benchmark results for mixed Snow\&Haze~\cite{chen2021all} dataset in PSNR (dB).
Our \textit{CoLoRA with PROD} achieved state-of-the art performance on NAFNet, by updating only a small 2M  of tuned network parameters.}
\begin{tabular}{c|c|c}
\toprule
Task                                   & Method                 & PSNR for mixed Snow\&Haze                      \\ \hline
\multirow{6}{*}{Independent methods}           & DesnowNet~\cite{liu2018desnownet}              & 25.63                     \\
                                       & JSTASR~\cite{chen2020jstasr}                 & 27.52                     \\
                                       & DesnowGAN~\cite{jaw2020desnowgan}              & 28.63                     \\
                                       & HDCW-Net~\cite{chen2021all}               & 29.11                     \\
                                       & DAD-S~\cite{zou2020deep}                  & 29.29 \\
                                       & MPRNet-S~\cite{zamir2021multi}               & 31.53 \\ \hline
\multirow{3}{*}{All-In-One methods}            & Chen~\cite{chen2022learning}                    & 32.28                     \\
                                       & Li~\cite{li2022all}                 & 26.91                     \\
                                       & Park~\cite{park2023all}                   & 32.41                     \\ 
                                       & PromptIR~\cite{potlapalli2023promptir}                   & 29.81                     \\ 
                                       & WGWS-Net~\cite{zhu2023Weather}                   & 29.16                     \\ 
                                       \hline
Efficient Fine-tuning & Our CoLoRA (NAFNet)    & 33.09                     \\
%                                       & Our CoLoRA (Restormer) & \multicolumn{1}{l}{}     
\bottomrule
\end{tabular}
    \label{supp:snow}
\end{table}

\section{Comparison of tuning strategies in DeGAE}
\begin{figure*}[t]
    \centering
    \includegraphics[width=1.0\textwidth]{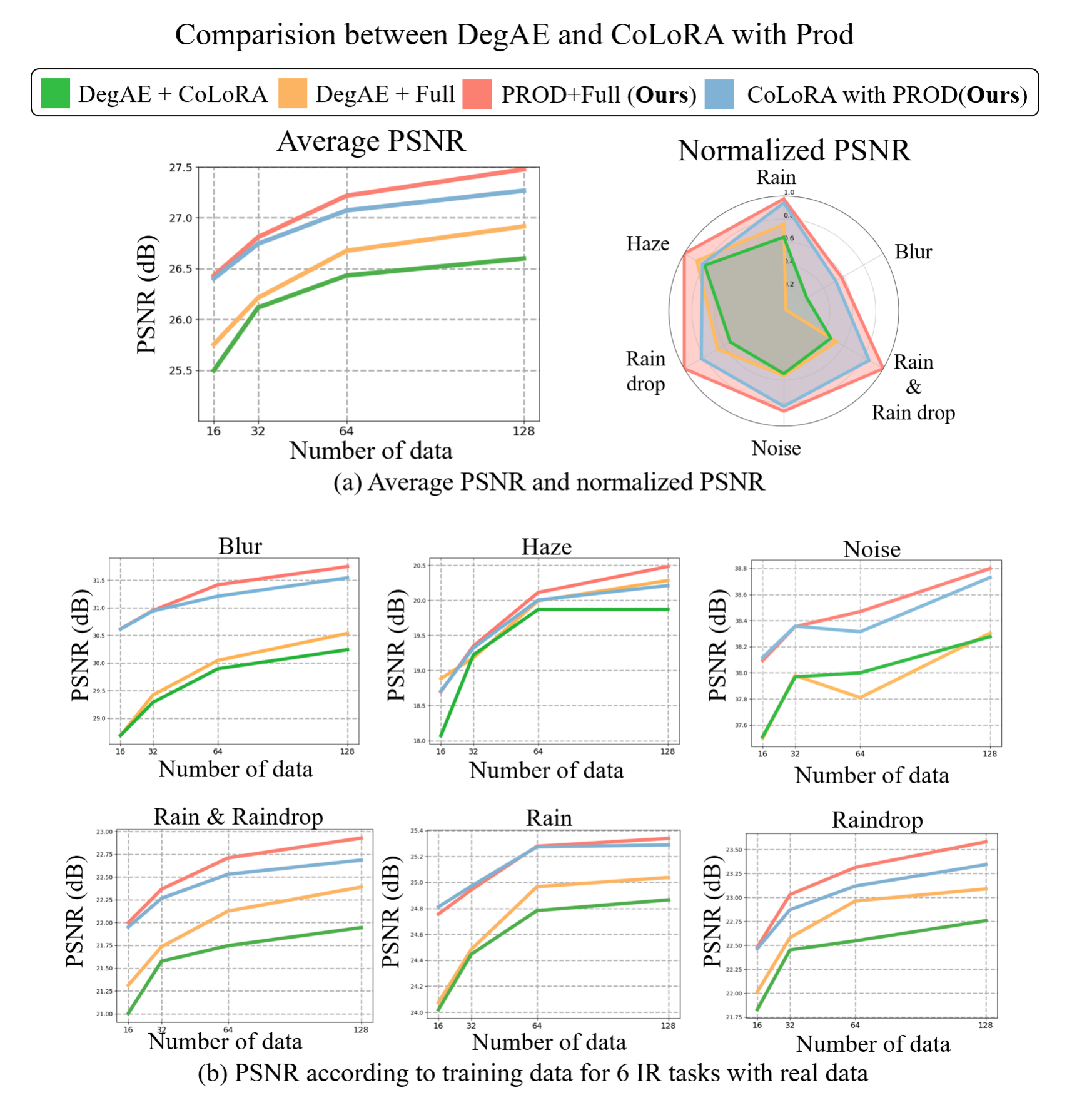}
     \vspace{-1.5em}
    \caption{
    Performance comparison of applying CoLoRA to DegAE method according to number of training data for 6 IR tasks.
    In the graph (a), the results of the 6 IR tasks are averaged for comparison.
    The x-axis represents the number of training data, and the y-axis is the average PSNR in the RGB domain.
    In the radar graph, we compare the results of 6 IR tasks with Normalized PSNR at a training data size of 128.
    In (b), the PSNR in the RGB domain is reported for each of the six tasks.
    The x-axis represents the number of training data, and the y-axis is the PSNR in the RGB domain for each test data. 
}
    \label{supp:degae}
        \vspace{1.3em}
\end{figure*}

In this section, we evaluated the proposed CoLoRA on 6 IR datasets with DegAE and a training iteration is 10K.
Figure~\ref{supp:degae} summarizes the experimental results based on DegAE with CoLoRA. 
All experiments are conducted using the CNN-based NAFNet architecture.
We investigated the following tuning methods: Full fine-tuning (\textit{Full}), and the proposed \textit{CoLoRA} method.
The \textit{Full} method has 29 M trainable parameters. \textit{CoLoRA}(Ours) have approximately 2 M trainable parameters.
Our proposed ``CoLoRA + DegAE'' method produces results similar to ``Full + DeGAE'' while tuning only very small network parameters.

%29159715
%2063651
\section{Details on comparison of pre-train methods and tuning strategies}
\begin{figure*}[t]
    \centering
    \includegraphics[width=0.95\textwidth]{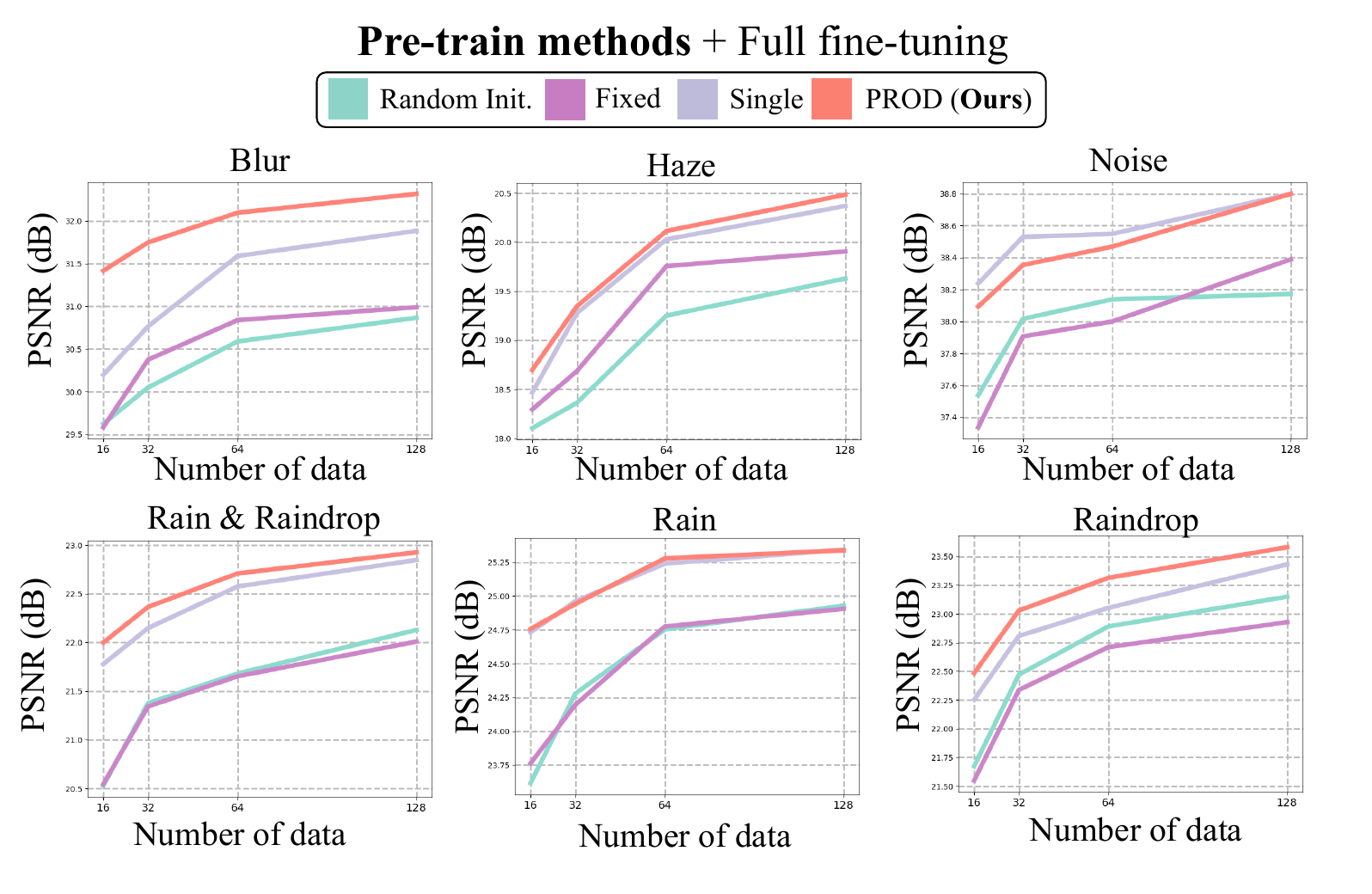}
    \includegraphics[width=0.95\textwidth]{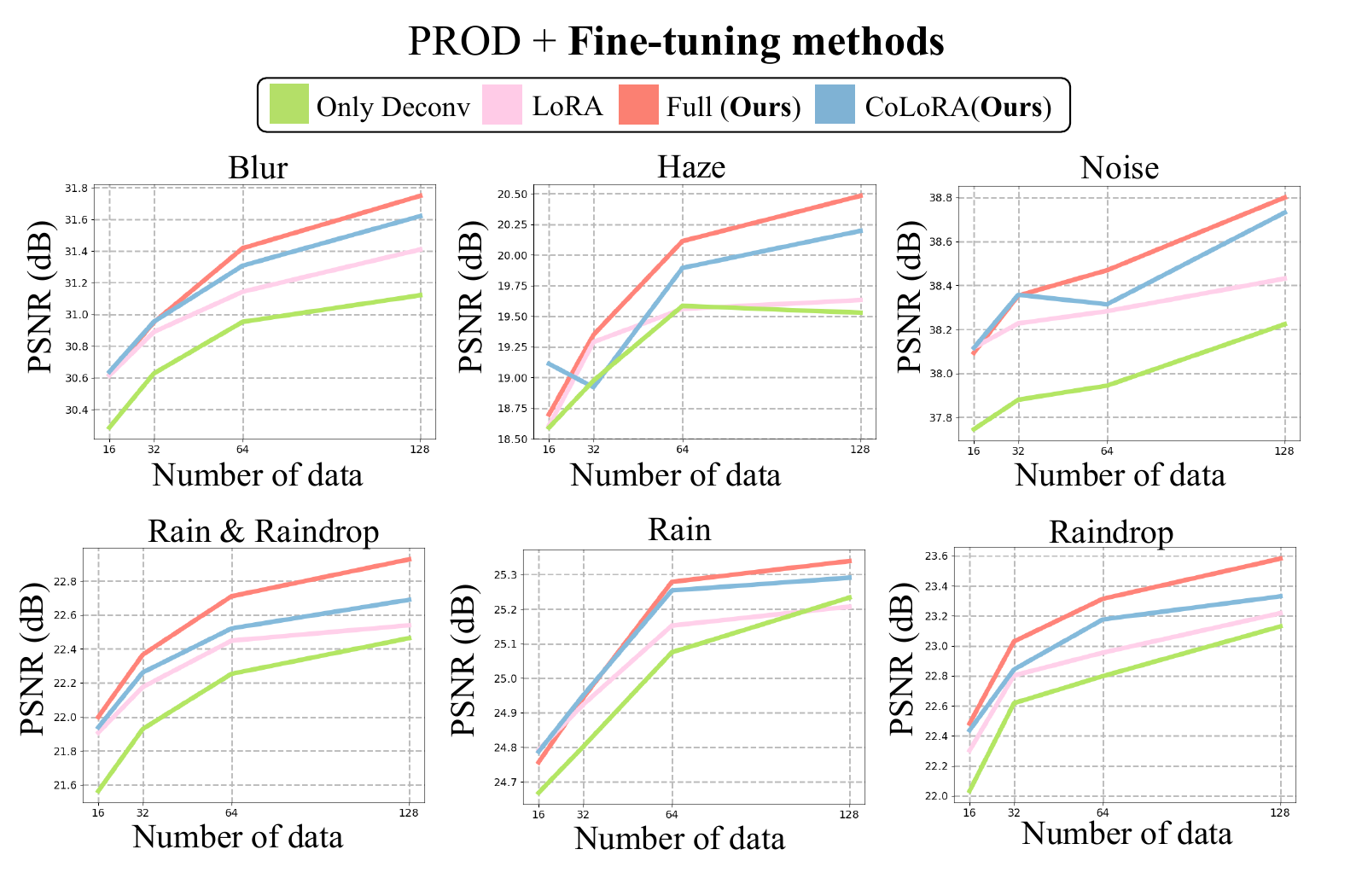}
     \vspace{0.5em}
    \caption{In section 4.1 of the paper, `pre-training methods' and `fine-tuning strategies', we conducted a performance comparison based on the scale of training data for 6 IR tasks with real data.
    The PSNR in the RGB domain is reported for each of the 6 tasks in Figure 5 of the paper.
    The x-axis represents the number of training data, and the y-axis is the PSNR (dB) for each test data. 
}
    \label{supp:pretrain}
        \vspace{-1em}
\end{figure*}

\begin{figure*}[t]
    \centering
    \includegraphics[width=0.9\textwidth]{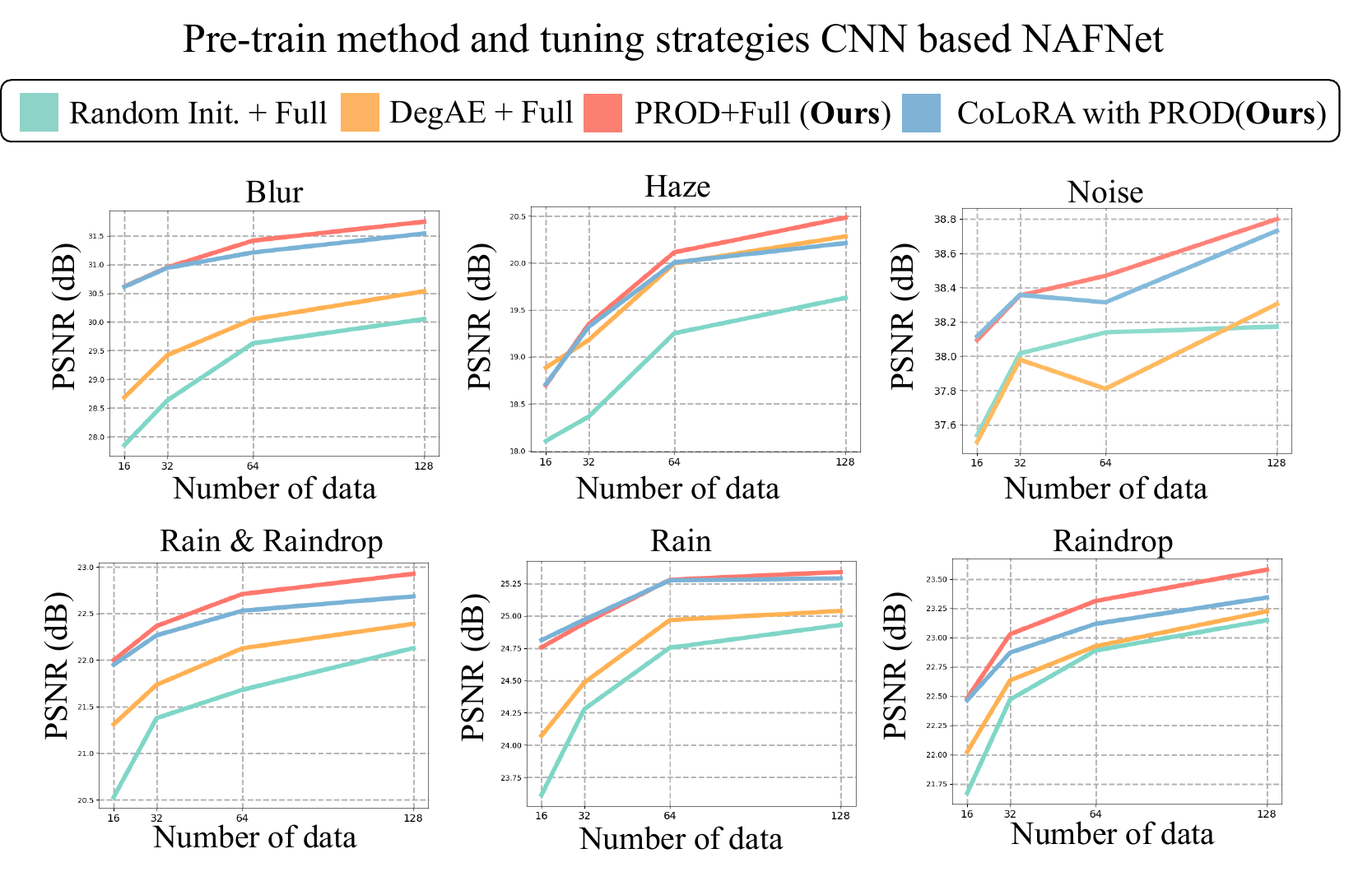}
    \includegraphics[width=0.9\textwidth]{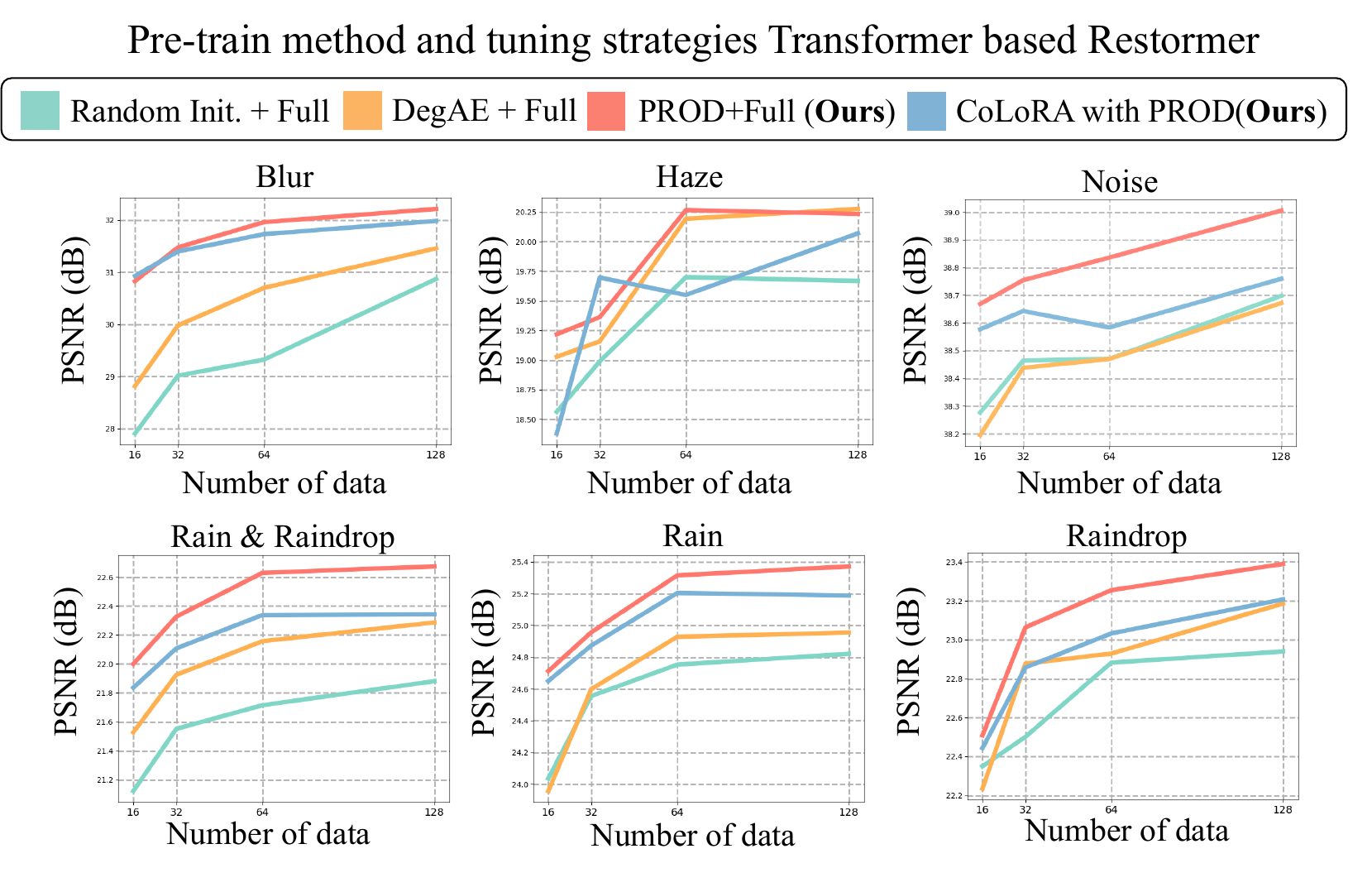}
     \vspace{0.5em}
    \caption{In section 4.1 of the paper, `different network architectures', we conducted a performance comparison based on the scale of training data for 6 IR tasks with \emph{real} data.
    The PSNR in the RGB domain is reported for each of the six tasks in Figure 5 of the paper.
    The x-axis represents the number of training data, and the y-axis is the PSNR for each test data. 
}
    \label{supp:network}
        \vspace{-1em}
\end{figure*}

We conducted a detailed comparative study to demonstrate the performance and efficiency of the proposed pre-training method PROD, and the parameter-efficient fine-tuning method, CoLoRA.
To summarize each experiment, ``pre-training methods'' presents the experimental results based on different pre-training methods in Figure~\ref{supp:pretrain} (top), ``tuning strategies'' presents the results based on efficient parameter tuning methods in Figure~\ref{supp:pretrain} (bottom), and ``different network architectures'' provides the experimental results based on pre-training and tuning methods on CNN and vision transformer architectures in Figure~\ref{supp:network}.
Figure~\ref{supp:pretrain} and ~\ref{supp:network} illustrate a detailed summary of the PSNR (dB) results for 6 IR tasks with real data in relation to the number of training data during fine-tuning.
The number of training data for each tasks were set to [16, 32, 64, 128]. 
These results demonstrate that the our proposed \textit{PROD + Full} and \textit{PROD + CoLoRA} method effectively and efficiently improves various IR performances, even with limited training data and a sparse number of learnable parameters, underscoring its effectiveness.

\begin{figure*}[t]
    \centering
    \includegraphics[width=1.0\textwidth]{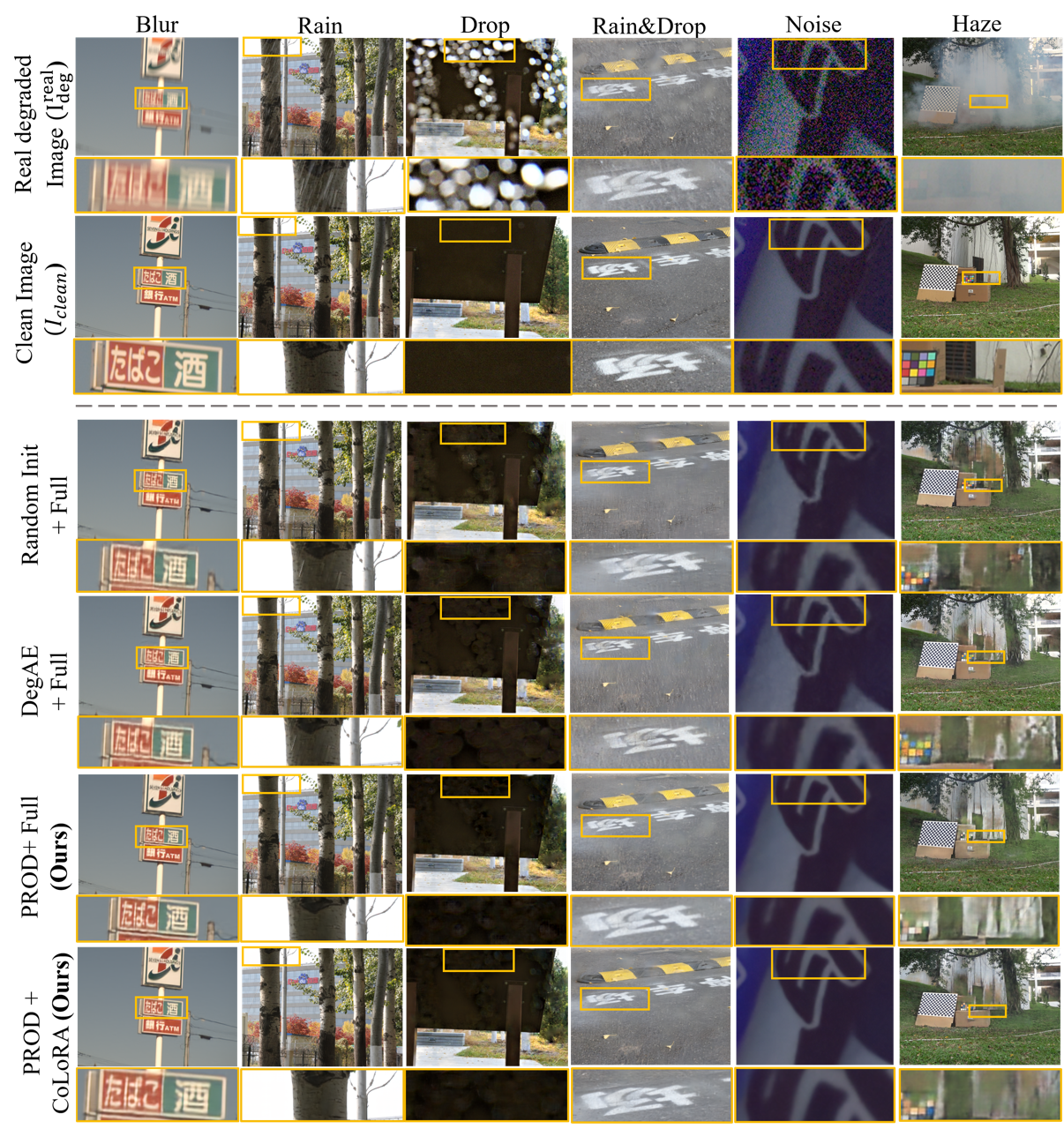}
    \caption{Qualitative results evaluated on the 6 IR tasks for our proposed CoLoRA with PROD, generic Random initial + Full tuning and DegAE + Full tuning. Our proposed methods with partial and full tuning yielded visually excellent results for the real IR tasks, outperforming others.}
    \label{supp:final_results_1}
        \vspace{0em}
\end{figure*}

\begin{figure*}[t]
    \centering
    \includegraphics[width=1.0\textwidth]{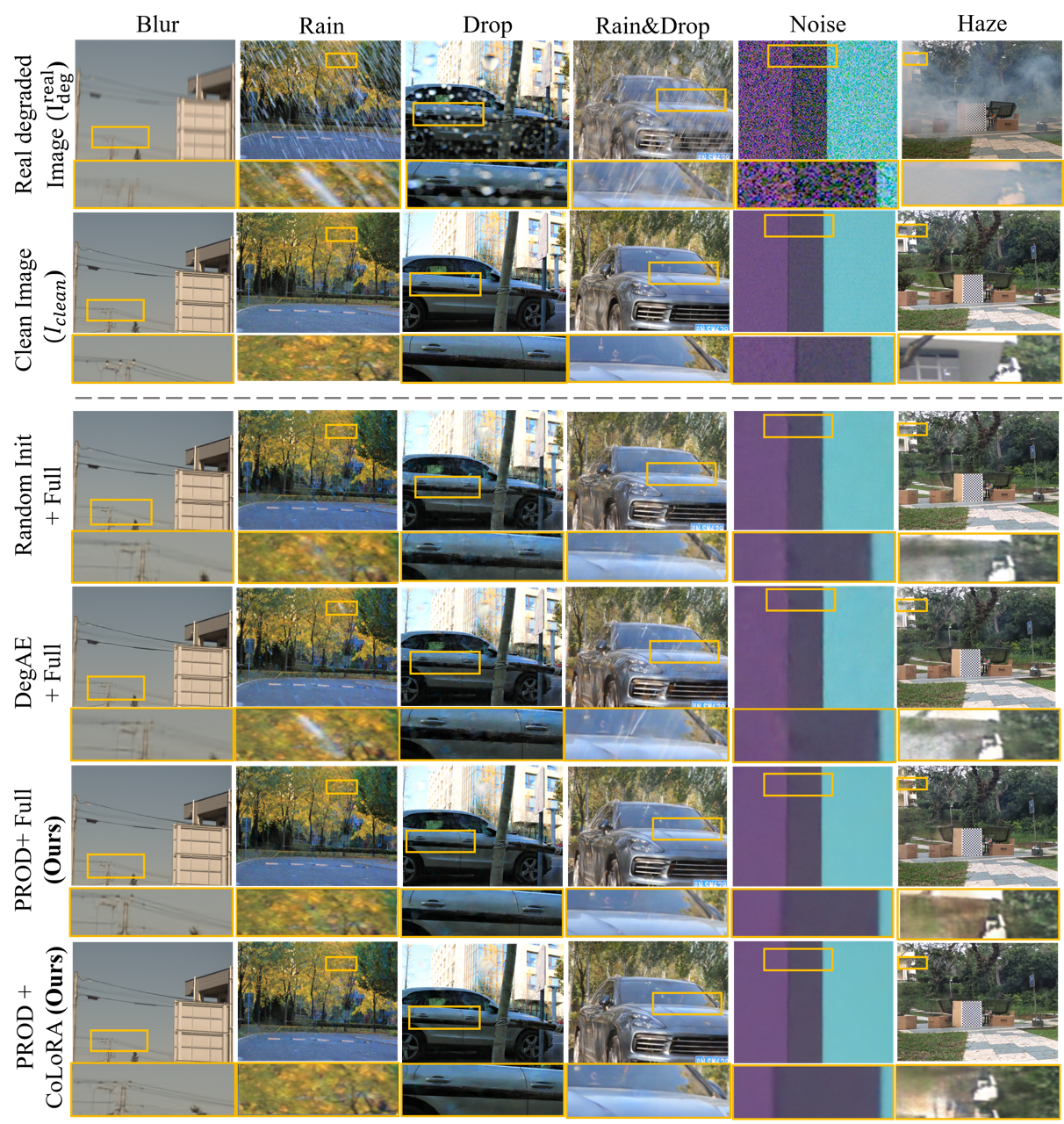}
    \caption{Qualitative results evaluated on the 6 IR tasks for our proposed CoLoRA with PROD, generic Random initial + Full tuning and DegAE + Full tuning. Our proposed methods with partial and full tuning yielded visually excellent results for the real IR tasks, outperforming others.}
    \label{supp:final_results_2}
        \vspace{0em}
\end{figure*}

\begin{figure*}[t]
    \centering
    \includegraphics[width=1.0\textwidth]{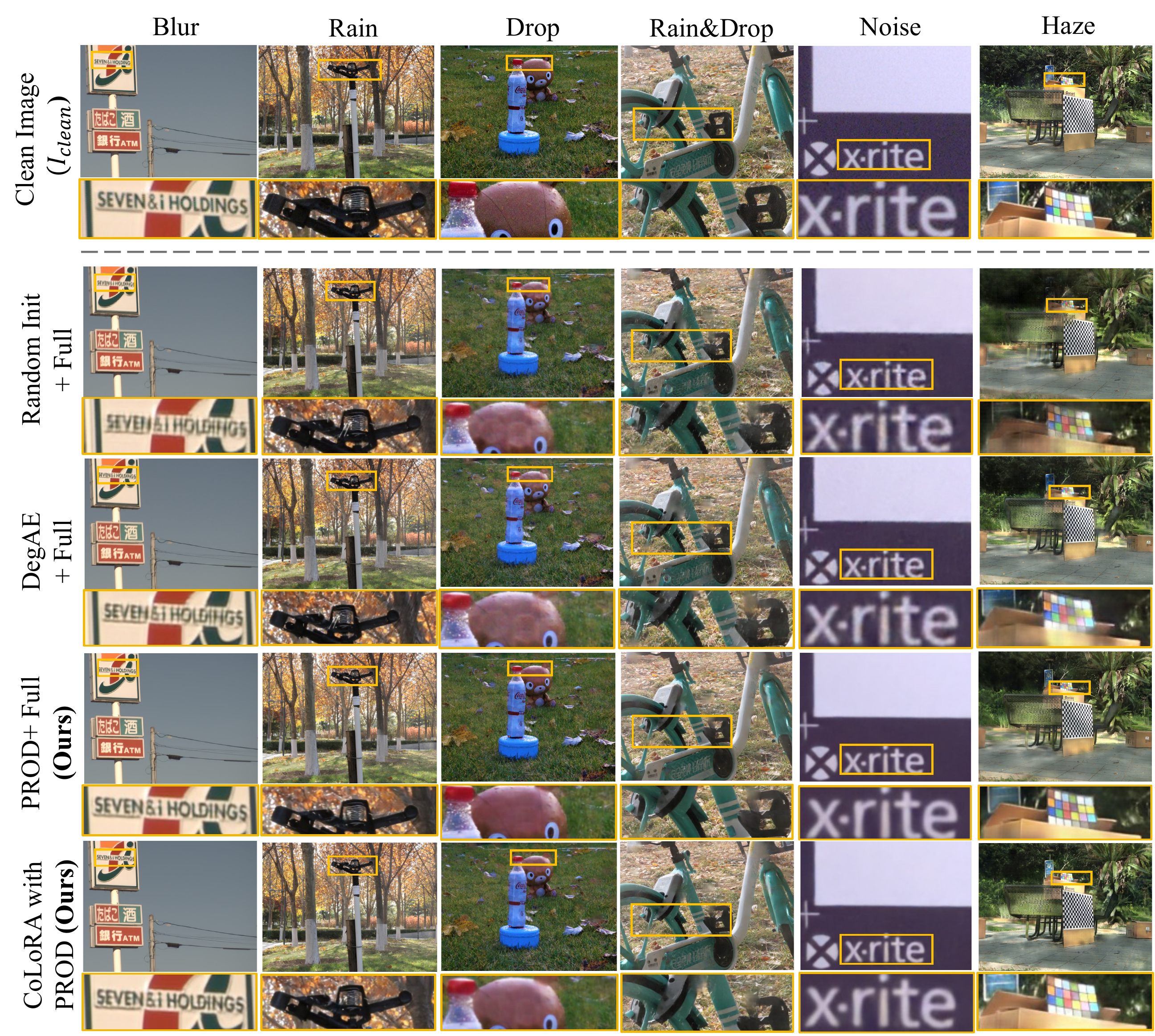}
    \caption{Qualitative results evaluated on the 6 IR tasks for our proposed CoLoRA with PROD, generic Random initial + Full tuning and DegAE + Full tuning. Our proposed methods with partial and full tuning yielded visually excellent results for the real IR tasks, outperforming others.}
    \label{supp:final_results_3}
        \vspace{0em}
\end{figure*}

Figure~\ref{supp:final_results_1}, Figure~\ref{supp:final_results_2} and~\ref{supp:final_results_3} illustrates the results for  6 IR task with test data tasks using test data, comparing outcomes between previous methods and our proposed CoLoRA with PROD. 
Our methods (\textit{PROD + Full} and \textit{PROD + CoLoRA}) appear to produce superior restoration results compared to \textit{Random + Full} and \textit{DegAE + Full}~\cite{liu2023degae}.


\end{document}